\documentclass[11pt]{article}

\usepackage[final]{acl}

\usepackage{times}
\usepackage{listings}
\usepackage{latexsym}
\usepackage{hyperref}
\usepackage{booktabs} 
\usepackage{float}
\usepackage[T1]{fontenc}
\usepackage[table]{xcolor}

\usepackage[utf8]{inputenc}

\usepackage{microtype}

\usepackage{inconsolata}

\usepackage{graphicx}
\usepackage{multirow}
\usepackage{comment}
\title{TailNLG: A Multilingual Benchmark Addressing Verbalization of Long-Tail Entities}

\author{
 \textbf{Lia Draetta\textsuperscript{1}},
 \textbf{Michael Oliverio\textsuperscript{1}},
 \textbf{Virginia Ram\'on-Ferrer\textsuperscript{2}},
 \textbf{Pier Felice Balestrucci\textsuperscript{1}},
 \\
 \textbf{Flaviana Corallo\textsuperscript{1}},
  \textbf{Carlos Badenes-Olmedo\textsuperscript{2}},
  \textbf{Alessandro Mazzei\textsuperscript{1}},
\\
\textbf{Marco Antonio Stranisci\textsuperscript{1}},
  \textbf{Rossana Damiano\textsuperscript{1}}
\\
\\
 \textsuperscript{1}University of Turin, Italy
 \textsuperscript{2}Universidad Politécnica de Madrid, Spain
   \\
  \small{
   \textbf{Correspondence:} \href{mailto:lia.draetta@unito.it}{lia.draetta@unito.it} 
  }
}

\begin{document}
\maketitle

\begin{abstract}
 Automatic verbalization of structured knowledge is a key task for making knowledge graphs accessible to non-expert users and supporting retrieval-augmented generation systems. Although recent advances in Data-to-Text generation have improved multilingual coverage, little attention has been paid to potential biases in the verbalization of rare entities, frequently known as long-tail entities. In this work, we present the first systematic study of long-tail entities in Data-to-Text generation. We introduce TailNLG\footnote{Full code and Benchmark available here \url{https://anonymous.4open.science/r/TailNLG-benchmark-B339/README.md}}, a new multilingual benchmark in English, Italian, and Spanish, built from Wikidata and covering entities with varying levels of popularity. We evaluated three different families of large language models in zero-shot settings and compared their performance on rare \textit{versus} common entities, as well as against the established WebNLG benchmark. Our results reveal a consistent bias against long-tail entities: embedding-based scores are lower, and model uncertainty is higher for rare entities. We further show that the impact of long-tail entities varies across models and languages and that existing evaluation metrics do not consistently capture these differences, highlighting the need for more reliable evaluation frameworks.
\end{abstract}

\section{Introduction}
\begin{figure}[h]
    \centering
    \includegraphics[width=1\linewidth]{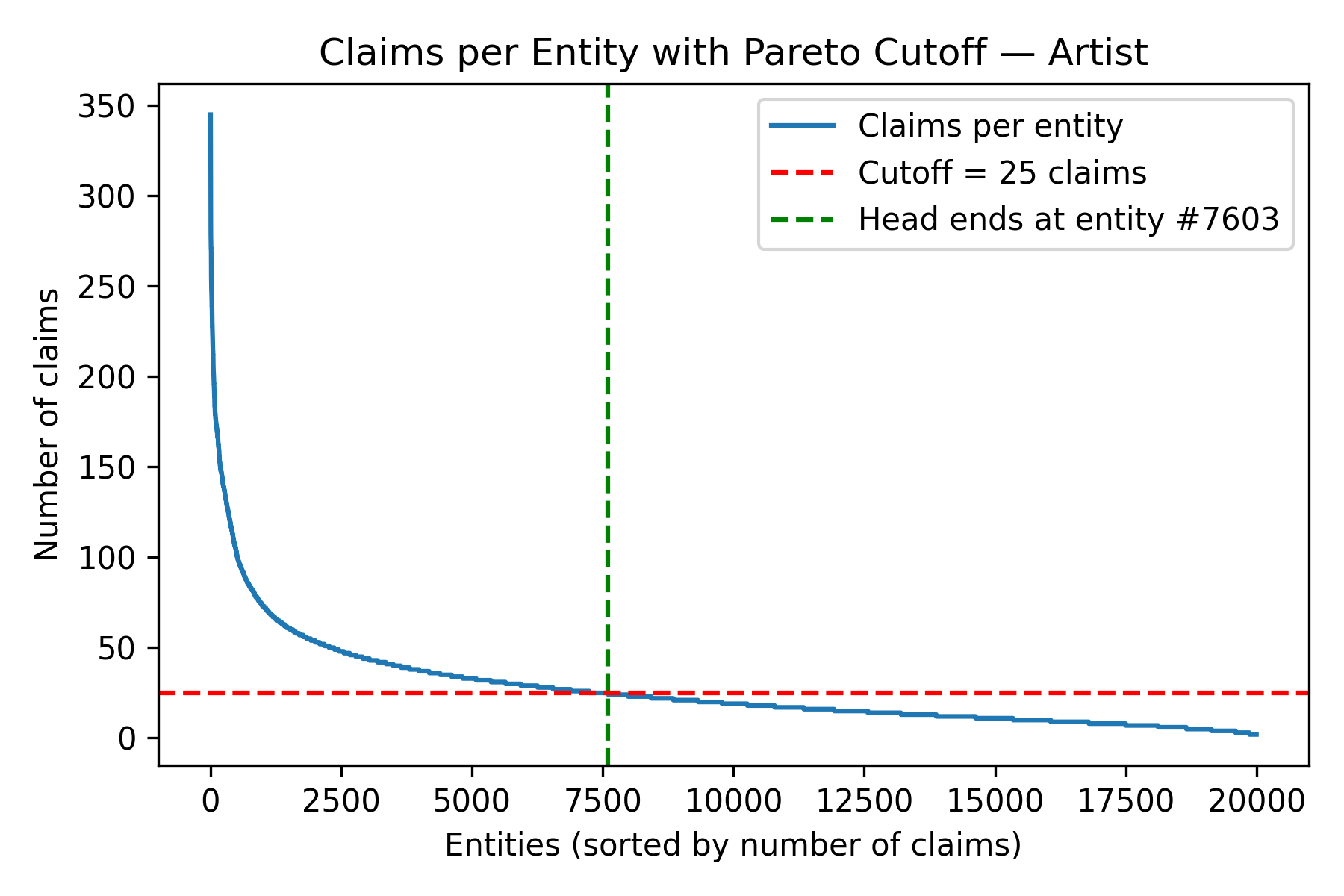}
    \caption{Distribution of entities by claims in the Wikidata category \textit{Artist}. A small number of entities are associated with many claims, while the majority (long-tail) are associated with few relations. The red dashed line indicates the Pareto cut-off, while the green dashed line marks the head–tail threshold}
    \label{fig:pareto_Artist}
\end{figure}

Automatic conversion of structured information into human-readable text is a crucial task for making knowledge accessible to a broad audience. The information contained in knowledge graphs (KGs), catalogues, and taxonomies is typically accessible only to domain experts, whereas the general public requires it to be expressed in natural language. This need also extends 
to 
retrieval-augmented generation (RAG) technologies, which leverage textual knowledge bases to improve the safety and accountability of large language models (LLMs).

Despite advancements in this field and the growing number of multilingual resources \cite{ferreira:18, shimorina:19, cripwell:23, oliverio2025webnlg, ramon2025spanish}, there is a lack of work aimed at uncovering biases in the verbalization of structured knowledge \citep{blodgett2020language}. Specifically, an open challenge is to gain a deeper understanding of how LLMs handle rare entities, frequently known as \textit{long-tail} entities, and whether their performance differs when compared to well-known entities. Although it has been demonstrated that LLMs exhibit performance drops when handling rare entities in tasks such as entity linking \cite{boscariol2024evaluation},  representation of factual knowledge \cite{sun-etal-2024-head}, and question answering \cite{hogan2025large}, no prior work has conducted an in-depth investigation of how different models handle long-tail entities in Data-to-Text generation. 

In this work, we tackle this gap by presenting the first systematic study of long-tail entities in Data-to-Text verbalization.
We focus on RDF-to-Text verbalization, as the Resource Description Framework (RDF) is a standard data model for representing information in Knowledge Graphs, encoding data as triples in the form ⟨subject, relation, object⟩.
Acknowledging the socio-cultural factors that can determine the rarity of an entity \citep{adams2019counts,stranisci2023world}, we adopt a multilingual approach. Our study introduces \textbf{TailNLG}, a novel multilingual RDF-to-Text benchmark covering three languages (English, Italian, and Spanish). The dataset contains triples extracted from Wikidata \cite{vrandevcic2014wikidata}, including entities from different categories and spanning varying levels of popularity, from top-head (most well-known) to long-tail (less well-known) entities.

Through the creation of TailNLG, our work addresses the following research questions:

\begin{itemize}
    \item \textbf{[RQ1]} Do LLMs perform differently in the verbalization of rare entities, and does language impact the verbalization output?
    \item \textbf{[RQ2]} Is the model uncertainty higher when dealing with rare \textit{versus} common entities in the verbalization task, and how does model performance differ when comparing a long-tail benchmark (TailNLG) with a well-established resource such as WebNLG?
\end{itemize}

To this end, three families of LLMs are assessed on the verbalization of long-tail entities in two settings: (i) comparing them with head entities within the TailNLG corpus, and (ii) comparing them with WebNLG \cite{gardent2017webnlg}, the established benchmark for the RDF-to-Text task. 
We assess performance with a diverse set of automatic measures, capturing semantic similarity, surface-level overlap with reference texts, and Perplexity (PPL) \cite{jelinek1977perplexity} to quantify model uncertainty.
%
Our results show that LLMs have a systematic bias against long-tail entities: embedding-based metrics tend to be 
lower for long-tail entities and models' uncertainty in text generation of triple containing long-tail entities is almost always higher. 
In addition, language-specific bias also plays a role in the verbalization of long-tail entities. Analysis of PPL scores reveals that embedding-based and overlap-based metrics fail to systematically quantify their impact on verbalization, demonstrating that a more reliable evaluation framework of fairness in RDF-to-Text is needed. 

\section{Related Work}
\label{sec:related works}

Long-tail entities are generally understood as those that occur with low frequency across large scale data sources, including training corpora \cite{kandpal2023large}, Wikipedia \cite{mallen2023not}, and large knowledge bases such as Wikidata \cite{kumar2024automatic}. 
Recent work has shown that LLMs exhibit a marked decline in performance when processing long-tail inputs  \cite{graciotti2025ke, hogan2025large, li-etal-2024-search, sun-etal-2024-head} and that LLMs memorization is highly influenced by the frequency of information in the pretraining data \cite{kandpal2023large}. 
Several task-specific evaluations further confirm this pattern.
\citet{boscariol2024evaluation} find that entity linking is particularly difficult for rare entities when comparing various LLMs with ReLiK \cite{orlando2024relik}, a state-of-the-art (SoTA) entity linking and relation extraction system. GRADES, an evaluation framework for graph-based question answering \cite{draetta2025beyond}, shows that SoTA models struggle with rare entities at multiple stages; in particular, verbalization suffers from limited semantic understanding of long-tail entities. In knowledge extraction, \citet{graciotti2025ke} introduce KEMISTO, a benchmark centered on low-popularity entities, and demonstrate that LLMs perform significantly worse on it and exhibit systematic biased failures.
A similar pattern emerges in question answering: using a benchmark spanning head, torso, and tail entities, \citet{sun-etal-2024-head} show that accuracy consistently declines as entity frequency decreases, confirming that limited training exposure hampers LLMs'  knowledge of long-tail entities.

Despite increasing attention to long-tail evaluation, no existing work specifically examines how well LLMs verbalize information when required to handle rare entities.
\\
\\
In the context of Data-to-Text, the task of verbalizing RDF triples in NLP has evolved substantially, driven both by template-based NLG systems and more recent neural approaches. Early work focused on hand-crafted or automatically induced templates, which provided high precision and strong control over the output,
but suffered from limited scalability and reduced the ability to generalize beyond predefined patterns \cite{duma2013generating}.
With the advent of neural encoder–decoder architectures, particularly LSTM-based triple encoders and later transformer models, RDF-to-Text generation became considerably more flexible, enabling fluent and information-rich verbalizations that better capture lexical and syntactic variation \cite{distiawan2018gtr, oliverio2024dipinfo}. Nevertheless, challenges related to factual consistency and robustness to unseen entities have motivated a series of enhancements to transformer-based systems. For example, integrating 
reward signals derived from information-extraction models has been shown to improve factual accuracy \cite{gao2021rdf}, while leveraging large external corpora, pre-training noise processes, and data augmentation strategies leads to more resilient generalization, particularly in low-resource or zero-shot scenarios \cite{montella2020denoising, zhang2023enhancing}.

To support systematic development and evaluation of this task, the WebNLG benchmark has emerged as one of the primary datasets for RDF-to-Text, providing multiple versions, multilingual extensions, and low-resource configurations that have enabled extensive comparative studies across approaches and languages \cite{gardent2017webnlg, webnlg-2020-international}.
\section{The TailNLG Benchmark}
\label{sec:benchmark}

Since no benchmark currently targets long-tail triple–verbalization pairs, we introduce \textsc{TailNLG}, the first dataset specifically designed for long-tail verbalization. 
Inspired by \citet{gardent2017webnlg}, we created the TailNLG benchmark following the same procedure employed by the authors,
from the entity extraction strategy to verbalization, with the aim of facilitating the use of data by adhering to a standard structure.
Constructing the benchmark in three languages adds another layer of complexity, as rare entities may not be consistently represented across multilingual resources. We prioritized high data quality by incorporating human evaluation and correction at each stage of the pipeline.

\subsection{Entities Selection}

To be consistent with previous resources, we followed the extraction methodology proposed by \citet{gardent2017creating} by selecting the same entity categories.
Although there is not yet a widely accepted and unambiguous definition of long-tail entities \cite{jiang-joshi-2024-cpopqa,boscariol2024evaluation,kumar2024automatic}, following \citet{sun-etal-2024-head} we took into consideration two main criteria in defining long-tail entities: (i) the number of claims (Wikidata relations) linked to a certain entity and (ii) the number of Wikipedia pages an entity has. 
Firstly, we set a threshold based on the number of claims linked to each entity. As noted by \citet{kumar2024automatic}, large-scale resources such as Wikidata typically follow a Zipf-like distribution, where a small number of entities are associated with many claims while the vast majority with few. 
Based on a subset of $20,000$ entities for each category (complete list of categories in Table \ref{tab:category_stats} in Appendix \ref{sec:categories}) we define the long-tail threshold using a Pareto cut-off: we compute the minimum number of claims required for an entity to fall within the top 20\% of entities that account for 80\% of all claims (the head).

Entities below this threshold are classified as long-tail. Since claim distributions differ across categories, we compute the cut-off separately for each category.
Figure \ref{fig:pareto_Artist} shows an example of the threshold for the categories Artists: a small number of entities exhibit a very high number of claims, while the vast majority fall into a long-tail distribution. 
To further refine the set of long-tail entities below the Pareto cut-off, we additionally select those that have a Wikipedia page only in one of the three target languages (e.g., entities with Wikipedia page in English but not in Italian and Spanish).

\begin{figure}[]
    \centering
    \includegraphics[width=1\linewidth]{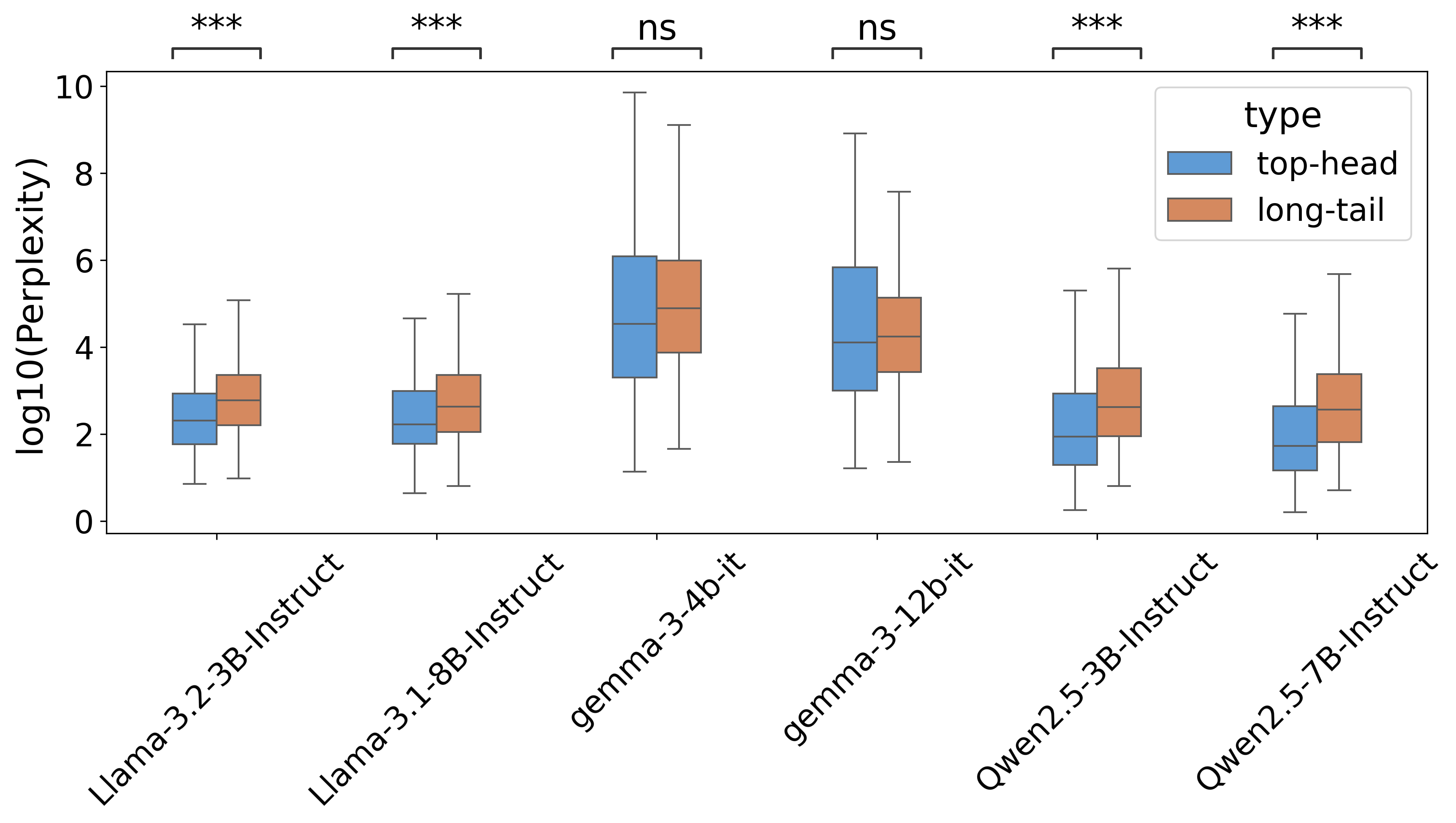}
    \caption{PPL distributions for head and long-tail entities across models (log10 scale).  Boxes represent the interquartile range (IQR, Q1–Q3) with the median shown as a line inside the box. Statistical significance was assessed using the \textit{Mann-Whitney U test}: *** p < 0.001, ns = not significant.}
    \label{fig:ppl_entities}
\end{figure}

In summary, we categorize entities into: (i) \textbf{long-tail entities}, whose number of claims falls below the Pareto cut-off, such as \textit{Doğan Güzel} (Q268285), a Turkish cartoonist; and (ii) \textbf{Head entities}, whose number of claims exceeds the Pareto cut-off, such as \textit{London} (Q84) or \textit{Google} (Q95).
\\

To examine model confidence on long-tail \textit{vs.} head entities, assessing the reliability of the TailNLG dataset, we compute the average Perplexity (PPL) as a proxy for model uncertainty across LLMs of different families and sizes (see Section~\ref{sec:methodology}). For each model, the entity string is tokenized using the model’s corresponding tokenizer, and PPL is computed as the exponential of the average negative log-likelihood of the resulting token sequence. Results in Figure~\ref{fig:ppl_entities} show that long-tail entities exhibit higher PPL than head entities, indicating greater model uncertainty when processing long-tail entities. This trend supports the reliability of our entity extraction methodology and the overall quality of the dataset. Details of the PPL calculations are provided in Appendix~\ref{sec:entity-ppl-analysis}.

\subsection{Triples Extraction}
The next step was to extract data units for each entity. 
We define a data unit as a set of RDF triples, each consisting of a subject, a predicate, and an object (e.g., Jet Li, place of birth, Beijing). To obtain diverse data units that vary in relation type and size, we followed \citet{perez-beltrachini-etal-2016-building} and extracted units with different relational configurations: Chain where the object of one triple serves as the subject of another, Sibling where the triples share the same subject and mixed where the data unit contains both sibling and chain relations. Some examples in appendix \ref{app:configurations}.


For each QID (Wikidata entity ID), we automatically extracted from Wikidata, via a SPARQL query, a set of $10$ data units per entity in the three languages. Since not all triples were available in all three languages, we further filtered the data units to ensure the alignment across languages, resulting in a set of parallel data units as exemplified below:

\begin{quote}
\footnotesize
\textbf{EN:} \texttt{[Solar, creator, Paul S. Newman], [Paul S. Newman, place of birth, Manhattan], [Paul S. Newman, award received, Inkpot Award]}\\
\textbf{ES:} \texttt{[Solar, creator, Paul S. Newman], [Paul S. Newman, lugar de nacimiento, Manhattan], [Paul S. Newman, premio recibido, Premio Inkpot]} \\
\textbf{IT:} \texttt{[Solar, creatore, Paul S. Newman], [Paul S. Newman, luogo di nascita, Manhattan], [Paul S. Newman, premio ricevuto, Premio Inkpot]}
\end{quote}

We obtained $1878$ data units per language, divided into $1224$ long-tail and $654$ head entities. The complete distribution per category and type is  in Appendix \ref{sec:categories}.

\subsection{Verbalization}
To build the benchmark, we aimed to produce <\textit{Data\_Unit, Verbalization}> pairs for all data in all languages. Because manual verbalization is time consuming we adopted a two-step approach: we first manually verbalized a subset of data units for each language, then cross-translated them into the remaining languages using SoTA machine translation models in a human-in-the-loop setup.

The verbalization process was carried out on a subset of $689$ data units by nine volunteer native speakers in the three languages. Annotators were proficient in at least two languages, allowing for a cross-check validation. They were instructed to verbalize the triples as naturally as possible, producing grammatically correct and fluent sentences while following guidelines designed to promote naturalness and ensure data homogeneity. All guidelines and some example of verbalization are reported in Appendix \ref{sec:verbalizion guidelines}. Annotators could skip a data unit if they judged it incorrect or incomplete. After verbalization, a validation phase followed in which the annotators cross-checked each other's output, confirming or revising the initial sentences (full guidelines in Appendix \ref{sec:validation}). 
Ambiguous cases were discussed collectively before a final decision was made. Following this verbalization and cross-check process, $73$ data units were excluded from the dataset.

\subsection{Translation}

The manual verbalization phase produced a total of $616$ gold verbalizations ($130$ in English, $142$ in Spanish, and $344$ in Italian\footnote{This linguistic imbalance is due to the varying availability of expert native-speaking annotators.}).
To create a fully parallel dataset  we adopted an approach based on Machine Translation (MT) and Automatic Post-Editing (APE), as proposed by  \citet{aditya-hari-etal-2023-webnlg}.
Drawing on earlier studies on cross-lingual extensions of triples-to-Text datasets \cite{oliverio2025webnlg, ramon2025spanish}, we leveraged DeepL for Automatic Translation\footnote{\url{https://www.deepl.com/translator}}. 
For each manually created verbalization in a source language (X) corresponding translations were automatically generated in the other two target languages (Y and Z), resulting in a total of $1{,}232$ automatic generated translations.

To mitigate known issues in MT, such as omissions and hallucinations \cite{jm3}, we introduced an APE phase. 
To identify errors in the translations, we leveraged the 3.5B version of xCOMET \citet{guerreiro:24} \footnote{https://huggingface.com/Unbabel/XCOMET-XL}, a multilingual model for quality estimation and error span extraction.
For each input, the model outputs a sentence-level score (ranging from \textit{Excellent} to \textit{Weak}), as well as an error spans detected field, which identifies potential errors and their positions within the sentence.

Subsequently, for the error correction phase, we adopted xTOWER\footnote{https://huggingface.com/sardinelab/xTower13B}, a multilingual decoder-only model proposed by \citet{treviso:24}, designed to provide natural-language explanations of errors present in a translation and suggest a corrected version of the sentence (example of prompt and output in Appendix \ref{sec:error correction}). 
We perform error detection and correction on all the MT translated data; the resulting output is  a parallel dataset consisting of data units and verbalizations in Italian, Spanish and English as follows:

{\footnotesize
\begin{verbatim}

<lex quality="gold" lid="Id1" lang="en">
    Themisthocles was born in Turkey and 
    he was a citizen of the Classical Athens. 
    Themistocles's spouse is Archippe.
</lex>
<lex quality="silver" lid="Id2" lang="es">
    Temístocles nació en Turquía y fue 
    ciudadano de la Atenas clásica. 
    El cónyuge de Temístocles es Arquippe.
</lex>
<lex quality="silver" lid="Id3" lang="it">
    Temistocle è nato in Turchia ed era un 
    cittadino dell'Atene classica. 
    La consorte di Temistocle era Archippe.
</lex>

\end{verbatim}
}

To ensure the quality of the \textit{silver} dataset (produced during the MT and APE phases), we manually evaluated sentences that were assigned a xCOMET sentence-level score below \textit{good} (i.e., rated as \textit{moderate} or \textit{weak}). We sampled $50$ verbalizations for each source–target language pair and involved eight volunteer native speakers with high proficiency in at least one of the considered language pairs (e.g., English and Italian). Each source-target language pair was annotated by two different annotators following a widely adopted taxonomy for the evaluation of Accuracy and Fluency
\cite{lommel-etal-2014-using} (evaluation guidelines in Appendix \ref{sec:MT assessment}). In addition, we asked the annotators to propose a corrected version of the automatically translated verbalization, increasing the number of gold reference.
\begin{table}[]
    \centering
    \scriptsize
    \begin{tabular}{c|cc}
        \toprule
$\textbf{Source}\rightarrow\textbf{Target}$ & \textbf{Accuracy} & \textbf{Fluency} \\
        \midrule
        $es\rightarrow en$ & $4.45\pm0.58$ & $4.99\pm0.07$\\
        $en \rightarrow es$ & $4.57\pm0.53$ & $4.95\pm0.15$\\
        $it \rightarrow en$ & $4.71\pm0.49$ & $4.89\pm0.32$\\
        $en \rightarrow it$ & $4.72\pm0.48$ & $4.97\pm0.12$\\
        $es \rightarrow it$ & $4.51\pm0.49$ & $4.91\pm0.22$\\
        $it \rightarrow es$ & $4.63\pm0.49$ & $4.93\pm0.20$\\
        \bottomrule
    \end{tabular}
    \caption{Average scores ($0$ to $5$) and standard deviation of the human evaluation on \textit{Accuracy} e \textit{Fluency}}
    \label{tab:manual evaluation scores}
\end{table}
In Table \ref{tab:manual evaluation scores}, the scores of the manual evaluation are reported. The annotators rated the sentences very highly across all language pairs, with Accuracy and Fluency consistently scoring above $4$ on a scale from $0$ to $5$ (Inter-annotator agreement is reported in Appendix~\ref{sec:Manual assessment of translation}). The high scores attributed to the silver standard suggest that the translated instances were considered reliable by evaluators, confirming the high quality of data, especially considering that the evaluation was conducted on inputs that received lower scores from xCOMET. 

After the translation and APE phases, we obtain a dataset of $1,845$ data units, consisting of $1,120$ Gold and $725$ Silver instances. The full distribution by language is reported in Table~\ref{tab:verb_distribution} in Appendix~\ref{sec:Gold silver distribution}.

\section{Methodology}
\label{sec:methodology}
To study multilingual RDF-to-Text verbalization, we leverage three families of open state-of-the-art multilingual instruction-tuned models, each evaluated at two comparable sizes: Llama-3.2-3B-Instruct~\footnote{\url{https://huggingface.co/meta-llama/Llama-3.2-3B-Instruct}}, Llama-3.1-8B-Instruct\footnote{\url{https://huggingface.co/meta-llama/Llama-3.1-8B-Instruct}}, Qwen2.5-3B-Instruct\footnote{\url{https://huggingface.co/Qwen/Qwen2.5-3B-Instruct}}, Qwen2.5-7B-Instruct\footnote{\url{https://huggingface.co/Qwen/Qwen2.5-7B-Instruct}}, Gemma-3-4B-IT\footnote{\url{https://huggingface.co/google/gemma-3-4b-it}}, and Gemma-3-12B-IT\footnote{\url{https://huggingface.co/google/gemma-3-12b-it}}. We focus on these model families because they are multilingual, widely used, and well established in the literature, openly available, and offered in moderate parameter ranges that enable efficient and reproducible experimentation. 

\label{sec:methodology}

\subsection{Experimental Setup}
\label{sec:experimental-setup}

Given a set of RDF-style triples $(s,p,o)$, 
the system produces a single-paragraph natural language description in English, Spanish, or Italian. 
We run experiments on two datasets: (i) WebNLG, used as a baseline benchmark, and (ii) TailNLG, used to analyze performance on data unit containing long-tail entities. 
Since our goal is to study biases and examine how LLMs handle rare entities, we deliberately adopt a zero-shot approach to avoid influencing model outputs with external examples. Prompts are written English, Spanish, or Italian depending on the target verbalization. Full prompts are available in Appendix \ref{sec:prompts}.

For each input we generate three candidate verbalizations to capture decoding variability. We use stochastic decoding with temperature sampling (temperature $=0.7$), which increases output diversity compared to greedy decoding while keeping generations coherent. Generations are bounded by a fixed maximum length (max\_new\_tokens $=256$) to prevent overly long outputs and to standardize comparisons across models and settings. 
Experiments were conducted on a server with two Intel Xeon Gold 5418Y CPUs (48 physical cores each, 96 threads total), around 504 GB RAM, and two NVIDIA L40 GPUs with approximately 46 GB VRAM each.

\subsection{Evaluation}
\label{sec:evaluation}
In evaluating the results, we compare multiple metrics designed to capture three characteristics of the generated output: (i) semantic similarity (embedding-based metrics), (ii) textual similarity (overlap-based metrics), and (iii) model uncertainty in text generation\footnote{Evaluations are performed using Hugging Face packages}. We use BERTScore \cite{zhang2019bertscore} to evaluate semantic similarity, computing pairwise cosine similarity between candidate and reference texts with contextual embeddings. Specifically, we adopt the rescaled-with-baseline variant (BERTScore$_r$), using a multilingual BERT model\footnote{\url{https://huggingface.co/google-bert/bert-base-multilingual-cased}} to ensure comparability across languages. To evaluate textual similarity, we adopt: BLEU \cite{papineni2002bleu}, chrF++ \cite{popovic2017chrf++}, and three variants of ROUGE \cite{lin2004rouge}: ROUGE$_1$, which measures unigram (word-level) overlap, ROUGE$_2$, which measures bigram overlap and ROUGE$_L$, which captures sentence-level structural similarity by identifying the longest common subsequences. To quantify models uncertainty, we computed the average PPL interpreting lower PPL values as a proxy of the higher confidence of a model in text generation. 

To assess the statistical significance of each metric, we applied the Wilcoxon–Mann–Whitney test \citep{fagerland2009wilcoxon} on triple verbalization of the same model on different populations. We repeated the test ten times on random samples of $500$ predictions and averaged the results. 
\section{Results}
\label{sec:Result}
In this section, we present results comparing long-tail and head entities, as well as across languages (RQ1). We also examine the model uncertainty on verbalizing rare and well-known entities by comparing PPL scores on TailNLG and WebNLG (RQ2). 

\begin{table*}[]
\centering
\scriptsize
\begin{tabular}{ll  ccccccc}
\toprule
\textbf{Type} & \textbf{Lang} & \textbf{BERT$_{r}$}~$\uparrow$ & \textbf{BLEU}~$\uparrow$ & \textbf{chrF}~$\uparrow$ & \textbf{R-1}~$\uparrow$ & \textbf{R-2}~$\uparrow$ & \textbf{R-L}~$\uparrow$ & \textbf{PPL}~$\downarrow$ \\
\midrule
\midrule
& & & &\textbf{Llama-8B} \\

\midrule
\multirow{3}{*}{Long-tail} & en & \cellcolor[HTML]{F2F2F2}0.61 &\cellcolor[HTML]{F2F2F2}0.25 &\cellcolor[HTML]{F2F2F2}60.39 &\cellcolor[HTML]{F2F2F2}0.66 &\cellcolor[HTML]{F2F2F2}0.43 &\cellcolor[HTML]{F2F2F2}0.56 &\cellcolor[HTML]{F2F2F2}93.36 \\
& es & 0.58 & 0.17 & 53.30 & 0.58 & 0.35 & 0.49 & 222.16 \\
& it &\cellcolor[HTML]{F2F2F2}0.46 &\cellcolor[HTML]{F2F2F2}0.13 &\,\,\,\cellcolor[HTML]{F2F2F2}47.73\textsuperscript{*} &\,\,\,\cellcolor[HTML]{F2F2F2}0.49\textsuperscript{*} &\,\,\,\cellcolor[HTML]{F2F2F2}0.29\textsuperscript{*} &\,\,\,\cellcolor[HTML]{F2F2F2}0.43\textsuperscript{*} &\cellcolor[HTML]{F2F2F2}88.06 \\
\midrule
\multirow{3}{*}{Head} & en & \,\,\,\cellcolor[HTML]{F2F2F2}0.65\textsuperscript{*} & \cellcolor[HTML]{F2F2F2}0.26 &  \cellcolor[HTML]{F2F2F2}61.20 &  \cellcolor[HTML]{F2F2F2}0.67 & \cellcolor[HTML]{F2F2F2}0.44 &  \cellcolor[HTML]{F2F2F2}0.57 &  \,\,\,\cellcolor[HTML]{F2F2F2}59.03\textsuperscript{*} \\
& es & 0.60 & 0.19 & 53.23 & 0.58 & 0.36 & 0.48 & \,\,\,94.52\textsuperscript{*} \\
& it &  \cellcolor[HTML]{F2F2F2}0.45 &  \cellcolor[HTML]{F2F2F2}0.13 & \cellcolor[HTML]{F2F2F2}45.58 & \cellcolor[HTML]{F2F2F2}0.46 & \cellcolor[HTML]{F2F2F2}0.27 &  \cellcolor[HTML]{F2F2F2}0.40 & \,\,\,\cellcolor[HTML]{F2F2F2}68.84\textsuperscript{*} \\
\midrule
\midrule
& & & &\textbf{Gemma-12B} \\
\toprule
\multirow{3}{*}{Long-tail} & en &\cellcolor[HTML]{F2F2F2}0.63 &\cellcolor[HTML]{F2F2F2}0.26 &\cellcolor[HTML]{F2F2F2}61.65 &\cellcolor[HTML]{F2F2F2}0.67 &\cellcolor[HTML]{F2F2F2}0.44 &\cellcolor[HTML]{F2F2F2}0.58 &\,\,\,\cellcolor[HTML]{F2F2F2}330.02\textsuperscript{*} \\
& es & 0.62 & 0.21 & 58.80 & 0.62 & 0.40 & 0.53 & \,\,\,611.42\textsuperscript{*} \\
& it &\cellcolor[HTML]{F2F2F2}0.61 &\cellcolor[HTML]{F2F2F2}\,\,\,0.24\textsuperscript{*} &\,\,\,\cellcolor[HTML]{F2F2F2}60.68\textsuperscript{*} &\,\,\,\cellcolor[HTML]{F2F2F2}0.64\textsuperscript{*} &\,\,\,\cellcolor[HTML]{F2F2F2}0.42\textsuperscript{*} &\,\,\,\cellcolor[HTML]{F2F2F2}0.55\textsuperscript{*} &\,\,\,\cellcolor[HTML]{F2F2F2}425.27\textsuperscript{*} \\
\midrule
\multirow{3}{*}{Head} & en &\,\,\,\cellcolor[HTML]{F2F2F2}0.66\textsuperscript{*} &\cellcolor[HTML]{F2F2F2}0.27 &\cellcolor[HTML]{F2F2F2}61.72 &\cellcolor[HTML]{F2F2F2}0.68 &\cellcolor[HTML]{F2F2F2}0.45 &\cellcolor[HTML]{F2F2F2}0.57 &\cellcolor[HTML]{F2F2F2}397.46 \\
& es & \,\,\,0.64\textsuperscript{*} & 0.22 & 57.51 & 0.61 & 0.39 & 0.51 & 926.65 \\
& it &\cellcolor[HTML]{F2F2F2}0.61 &\cellcolor[HTML]{F2F2F2}0.20 &\cellcolor[HTML]{F2F2F2}56.97 &\cellcolor[HTML]{F2F2F2}0.59 &\cellcolor[HTML]{F2F2F2}0.38 &\cellcolor[HTML]{F2F2F2}0.50 &\cellcolor[HTML]{F2F2F2}1068.31 \\
\midrule
\midrule
& & & &\textbf{Qwen-7B} \\
\toprule
\multirow{3}{*}{Long-tail} & en &\cellcolor[HTML]{F2F2F2}0.65 &\cellcolor[HTML]{F2F2F2}0.29 &\cellcolor[HTML]{F2F2F2}62.47 &\cellcolor[HTML]{F2F2F2}0.69 &\cellcolor[HTML]{F2F2F2}0.46 &\cellcolor[HTML]{F2F2F2}0.58 &\cellcolor[HTML]{F2F2F2}134.13 \\
& es & 0.60 & 0.16 & 53.42 & 0.57 & 0.34 & 0.49 & 181.75 \\
& it &\cellcolor[HTML]{F2F2F2}0.44 &\cellcolor[HTML]{F2F2F2}0.11 &\cellcolor[HTML]{F2F2F2}46.82 &\cellcolor[HTML]{F2F2F2}0.47 &\cellcolor[HTML]{F2F2F2}0.26 &\cellcolor[HTML]{F2F2F2}0.41 &\cellcolor[HTML]{F2F2F2}100.26 \\
\midrule
\multirow{3}{*}{Head} & en &\,\,\,\cellcolor[HTML]{F2F2F2}0.69\textsuperscript{*} &\cellcolor[HTML]{F2F2F2}0.29 &\cellcolor[HTML]{F2F2F2}63.06 &\cellcolor[HTML]{F2F2F2}0.70 &\cellcolor[HTML]{F2F2F2}0.47 &\cellcolor[HTML]{F2F2F2}0.59 &\,\,\,\cellcolor[HTML]{F2F2F2}98.05\textsuperscript{*} \\
& es & 0.61 & 0.16 & 51.83 & 0.55 & 0.32 & 0.46 & \,\,\,102.36\textsuperscript{*} \\
& it &\cellcolor[HTML]{F2F2F2}0.46 &\cellcolor[HTML]{F2F2F2}0.10 &\cellcolor[HTML]{F2F2F2}45.35 &\cellcolor[HTML]{F2F2F2}0.45 &\cellcolor[HTML]{F2F2F2}0.25 &\cellcolor[HTML]{F2F2F2}0.39 &\,\,\,\cellcolor[HTML]{F2F2F2}63.48\textsuperscript{*} \\
\bottomrule
\end{tabular}
\caption{Results on TailNLG long-tail vs head entitities verbalizations by data type and language. Metrics include BERTScore Rescaled (BERT$_{r}$), BLEU,  chrF++ (chrF), ROUGE$_1$ (R-1) , ROUGE$_2$ (R2), ROUGE$_L$ (R-L) and Perplexity (PPL). Asterisks ($^{*}$) denote a statistically significant difference ($p < 0.05$).}
\label{tab:Results rq1}
\end{table*}

\subsection{[RQ1] Do LLMs perform differently in the verbalization of rare entities and does language impact the verbalization output?}

Table~\ref{tab:Results rq1} reports the scores for triples containing long-tail and head entities, broken down by language (English, Italian and Spanish), for the bigger size models. Complete results are provided in Appendix~\ref{sec:results appendix}. 
In all cases, the embedding-based metric BERTScore$_r$, scores higher on head triples. Conversely, results from chrF++, which correlates with morphological accuracy, and ROUGE are less consistent: in some cases they score higher on long-tail triples, while in others on head triples. 
In addition, for BERTScore$_r$, statistically significant results consistently show higher scores for head entities than for long-tail entities. In contrast, overlap-based metrics display the opposite pattern: when results are statistically significant, triples containing long-tail entities achieve higher scores. This difference may be due to the addition of lexical items when verbalizing triples with head entities, which can make the generated output diverge from the gold standard (see Section \ref{sec:Analysis}).
The average PPL provides clear insights, particularly for Qwen and Llama models. The results suggest that verbalizing triples with long-tail entities leads to higher PPL, suggesting that these entities present a substantially more challenging prediction task for the models. In contrast, the PPL scores for Gemma models are less clear and always lower for long-tail entities.
Investigating whether language influences the generated output, we observe that across all settings performance is higher for English, indicating that English content is handled more effectively by the models. In addition, for Llama and Qwen, average PPL scores for Italian are consistently higher than those for English and Spanish, indicating that the models could be affected by language bias and that outputs in more widely represented languages tend to be more accurate. \\
To better understand whether the presence of silver standard data influenced performance, we also split the results by gold \textit{vs.} silver (Tables \ref{tab:results_per_quality} and \ref{tab:multi_model_results} in the Appendix). Overall, the scores tend to be higher for the silver standard, indicating, in line with \cite{panickssery2024llm}, that the models are better aligned with machine-translated inputs. The comparison between Silver and Gold inputs also reveals interesting insights when results are further split into long-tail vs. head entities (Table \ref{tab:multi_model_results}): when the differences are statistically significant, scores for Silver data are higher for head entities, suggesting that machine translation and post-editing models may also exhibit a bias against less frequent entities. \\

\subsection{[RQ2] How does model performance differ when comparing the TailNLG benchmark with WebNLG?}

\begin{table}[t]
\centering
\scriptsize
\setlength{\tabcolsep}{3pt}
\setlength{\tabcolsep}{2pt}
\begin{tabular}{lccccccc}
\toprule
\textbf{Type} &
\textbf{BERT$_{r}$}~$\uparrow$ &
\textbf{BLEU}~$\uparrow$ &
\textbf{chrF}~$\uparrow$ &
\textbf{R-1}~$\uparrow$ &
\textbf{R-2}~$\uparrow$ &
\textbf{R-L}~$\uparrow$ &
\textbf{PPL}~$\downarrow$ \\
\midrule
\midrule
\multicolumn{8}{c}{\textbf{Llama-8B}} \\
\midrule
Long-tail & \cellcolor[HTML]{F2F2F2}0.55 & \cellcolor[HTML]{F2F2F2}0.18 & \cellcolor[HTML]{F2F2F2}53.81 & \cellcolor[HTML]{F2F2F2}0.58 & \cellcolor[HTML]{F2F2F2}0.36 & \cellcolor[HTML]{F2F2F2}0.50 & \cellcolor[HTML]{F2F2F2}134.53 \\
WebNLG    & 0.53 & \,\,\,0.21\textsuperscript{*} & 53.77 & \,\,\,0.65\textsuperscript{*}   & \,\,\,0.43\textsuperscript{*}   & \,\,\,0.53\textsuperscript{*}   & \,\,\,68.23\textsuperscript{*} \\
\midrule
\midrule
\multicolumn{8}{c}{\textbf{Gemma-12B}} \\
\midrule
Long-tail & \cellcolor[HTML]{F2F2F2}0.62 & \cellcolor[HTML]{F2F2F2}0.24 & \cellcolor[HTML]{F2F2F2}60.38 & \cellcolor[HTML]{F2F2F2}0.64 & \cellcolor[HTML]{F2F2F2}0.42 & \cellcolor[HTML]{F2F2F2}0.55 & \cellcolor[HTML]{F2F2F2}455.57 \\
WebNLG    & \,\,\,0.65\textsuperscript{*} & \,\,\,0.30\textsuperscript{*} & \,\,\,63.63\textsuperscript{*} & \,\,\,0.73\textsuperscript{*}   & \,\,\,0.50\textsuperscript{*}   & \,\,\,0.59\textsuperscript{*}   & \,\,\,290.10\textsuperscript{*} \\
\midrule
\midrule
\multicolumn{8}{c}{\textbf{Qwen-7B}} \\
\midrule
Long-tail & \cellcolor[HTML]{F2F2F2}0.56 & \cellcolor[HTML]{F2F2F2}0.19 & \cellcolor[HTML]{F2F2F2}54.24 & \cellcolor[HTML]{F2F2F2}0.58 & \cellcolor[HTML]{F2F2F2}0.35 & \cellcolor[HTML]{F2F2F2}0.49 & \cellcolor[HTML]{F2F2F2}138.71 \\
WebNLG    & 0.57 & \,\,\,0.24\textsuperscript{*} & 56.49 & \,\,\,0.66\textsuperscript{*}   & \,\,\,0.44\textsuperscript{*}   & \,\,\,0.54\textsuperscript{*}   & \,\,\,95.54\textsuperscript{*} \\
\bottomrule
\end{tabular}
\caption{Performance comparison of  models  between long-tail triples from TailNLG and WebNLG test set. Metrics: BERTScore Rescaled (BERT$_r$), BLEU, chrF++ (chrF), ROUGE-1 (R-1), ROUGE-2 (R-2), ROUGE-L (R-L), and Perplexity (PPL). Asterisks ($^{*}$) denote a statistically significant difference ($p < 0.05$).}

\label{tab:web_vs_tail}
\end{table}

Table~\ref{tab:web_vs_tail} reports the average performance of each model on the WebNLG test set and on long-tail triples from TailNLG. All statistically significant results indicate that embedding- and overlap-based metrics achieve higher scores on WebNLG than in the long-tail setting. This trend suggests that models generate more accurate and lexically aligned verbalizations for well-known entities.

Across all models, PPL is even lower for WebNLG verbalizations, with statistical significance, supporting the hypothesis that WebNLG entities and triples are more familiar to the models, potentially due to exposure during pre-training.

In addition, since models tend to achieve higher scores when synthetic data are used as ground truth \cite{sarti-nissim-2024-it5}, our findings are further reinforced. Even when comparing model performance on the English WebNLG test set (entirely gold) with the long-tail portion of TailNLG (containing both gold and silver references), models still achieve higher scores on WebNLG. For further details, we refer the reader to Appendix~\ref{sec:results appendix}.

\section{Discussion}
\label{sec:Analysis}

Results show that, across all models, verbalization of triples that contains long-tail entities is more challenging (RQ1). This suggests that LLMs suffer from entity bias, with widely represented entities being processed more easily and producing more accurate outputs.
The fact that in some cases overlap-based metrics are higher for data units containing long-tail entities suggest that the verbalization of head entities may contains more data and key content relative to the reference
For example, the Italian verbalization of the data unit \texttt{Karl Marx | spouse | Jenny von Westphalen; Karl Marx | country of citizenship | statelessness} is “Karl Marx è stato apatico alla cittadinanza, preferendo lo status di apolidia. Karl Marx ebbe come consorte Jenny von Westphalen” (English translation: Karl Marx was apathetic toward citizenship, preferring to remain stateless. Karl Marx’s spouse was Jenny von Westphalen), which results in a more verbose output that differs from the target sentence.
The results concerning language strongly confirm the hypothesis that models also suffer from language bias, and that verbalizations in English are more accurate. 
These results may also be influenced by the fact that both Italian and Spanish have richer morphologies than English. For example, the Italian verbalization of the triples \texttt{Amanda Ammann | country of citizenship | Switzerland; Amanda Ammann | occupation | model} results in "Amanda Ammann è una cittadina svizzera e lavora come modello" (English translation: Amanda Ammann is a Swiss citizen and works as a model), which exhibits a gender disagreement between the first part of the sentence (“cittadina svizzera”) and the final noun (“modello”). Generally Italian output tend to be less accurate and suffer from some additions and omissions.

Another important finding emerging from the results is the high variability across evaluation metrics. In particular, reference-based metrics are not always able to detect the presence of bias or provide reliable scores, as becomes evident from the comparison with PPL, which yields more consistent results. In addition, the analysis suggests that different models may be affected by different types of bias; in particular, Gemma exhibits a higher PPL for triples containing long-tail entities.
The use of PPL as proxy for uncertainty confirms our hypothesis that models perform better on a benchmark such as WebNLG compared to TailNLG (RQ2), as the models may have been exposed during pre-training to information related to the WebNLG benchmark or to the entities it contains. This finding is supported, with statistical significance, by both reference-based evaluation 
metrics and PPL.

The results break down by Silver \textit{vs} gold, also have practical implications for downstream applications that rely on verbalization,  showing that in pipelines where verbalized outputs are used as input to LLMs, silver-standard data may suffice, as models appear better aligned with machine-translated or automatically generated text. This suggests that silver data could provide usable outputs while reducing the need for fully curated gold-standard annotations, although further research is needed to evaluate their impact on model performance and reliability.

\section{Conclusion and Future Works}
\label{sec:Conclusion}

In this work, we investigated the performance of three state-of-the-art language models' families in depth for verbalizing long-tail entities in a multilingual setting. To this end, we introduced TailNLG, the first benchmark composed of triple–verbalization pairs covering entities that range from rare to well-represented. Our results show that the leveraged models suffer from both language and entity biases, and that they perform better on a widely exploited benchmark such as WebNLG compared to TailNLG.
Consistent with prior work, our findings highlight that current models struggle to handle long-tail entities, revealing limitations that make tasks involving less-represented entities more difficult and less reliable. 

As future work, we plan to enrich the TailNLG benchmark by adding more languages and data units. We also aim to refine the dataset by distinguishing entity types not only based on their popularity (i.e., number of claims) but also by incorporating culturally specific data units to investigate potential model biases. Additionally, we plan to create a gold standard that reflects diverse annotator perspectives, enabling the study of human label variation in the verbalization task.

\section*{Limitations}
We acknowledge that building a new high-quality benchmark is challenging, as it requires achieving an appropriate balance between human annotation and automatic processing steps. In this work, we consistently paired the automatic phase with human assessment in an effort to ensure the highest possible quality. Nevertheless, the presence of \textit{silver} verbalizations in the benchmark constitutes a limitation, since machine translation tools, although reliable, may not fully align with human judgments.

Second, by leveraging a well-known resource such as Wikidata, the variability of available claims is limited. As a result, models may already be familiar with many of the relations, which also recur across different data units.

Finally, in defining long-tail entities, we adopted a simplified approach for experimental purposes, considering only a quantitative parameter (number of claims). However, we are aware that entity underrepresentation is a multifaceted issue that involves multiple dimensions, both vertical (numerosity) and horizontal (cultural aspects). In future work, it would be valuable to consider additional perspectives and to define long-tail entities using a more comprehensive characterization.

\section*{Ethical Considerations}
Our work relies on knowledge extracted from Wikidata, whose community of contributors has recognized the underrepresentation of non-Western and woman contributors. This has an impact on how knowledge is shaped and which entities suffer underrepresentation. Adopting a purely statistical approach to the definition of \textit{long-tail} entities might have resulted in overlooking these cultural factors. Nevertheless, this approach was preferred to the adoption of theoretically grounded models of underrepresentation, as the latter could have introduced additional research bias in corpus creation. 

All annotations have been performed voluntarily by the members of our research group without relying on external annotation platforms.

During the preparation of this work, the authors used AI tools in order to perform Grammar and spelling check. After using this tool, the authors reviewed and edited the content as needed and take full responsibility for the publication’s content.


\bibliography{custom}

\appendix

\section{Entity Perplexity Analysis}
\label{sec:entity-ppl-analysis}

To compute PPL as a proxy for model uncertainty, we evaluated the six language models described in Section~\ref{sec:methodology}. This allows us to assess whether the PPL gap between head and long-tail entities is consistent across different model families. For each model, we tokenize the entity string directly using its specific tokenizer. PPL is calculated as the exponentiated average negative log-likelihood of the sequence. We performed a statistical comparison between head and long-tail distributions using the two-sided Mann-Whitney U test. Results are reported in Table~\ref{fig:ppl_entities} using median and interquartile range (IQR, Q1–Q3) and are visualized on a $\log_{10}$ scale to improve readability. In all settings, PPL values for long-tail entities are higher with statistical significance in most cases, thus validating our choice of using the Pareto cut-off for the selection of long-tail and top-head entities.

\section{Triples configuration}
\label{app:configurations}
\begin{figure}
    \centering
    \includegraphics[width=1\linewidth]{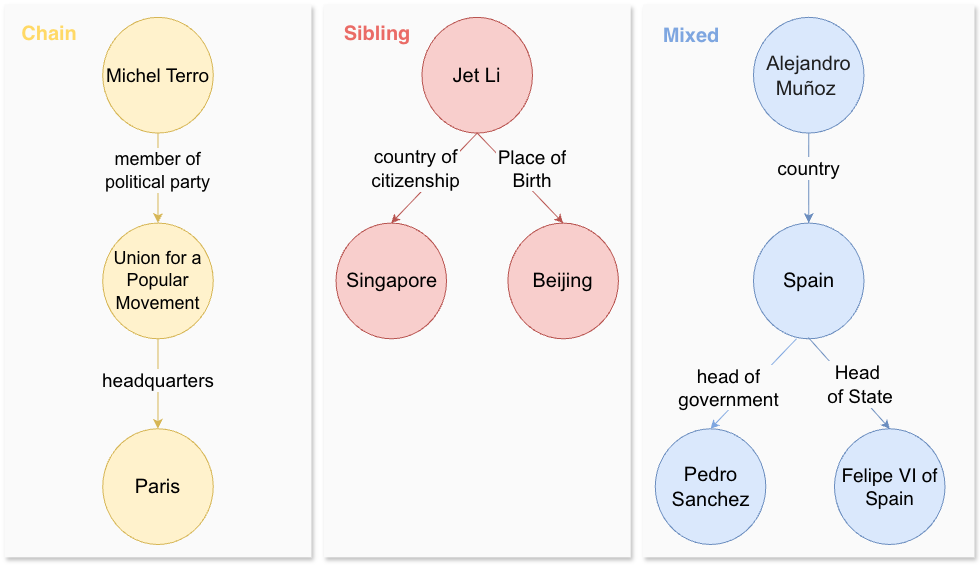}
    \caption{Examples of Chain, sibling and mixed triples configuration}
    \label{fig:configurations}
\end{figure}

In line with other resources in the field of NLG, we extracted entities using different configurations. Some examples are shown in Figure \ref{fig:configurations}.\ref{fig:configurations}

\section{Verbalization guidelines}

The annotators were asked to verbalize the data-units following guidelines:

\subsection{Verbalization}
\label{sec:verbalizion guidelines}

\begin{itemize}
    \item Verbalize the triples as naturally as possible, aiming to create grammatically correct and fluent sentences. You may choose whether or not to follow the original order of the triples. Please indicate this in the “Order preserved” column by writing YES if the triple is verbalized in the same order, or NO if it is not.
    \item You are free to break down the information conveyed by each set of triples in one or more sentences according to your preferences.
    \item If a triple makes no sense (e.g., "Black Racer", "different from", "Black Racer"), you may choose not to verbalize it. In this case, write NA in the annotation column.
    \item If a triple is very difficult or uncertain to interpret, you may propose a verbalization and leave comments in the Notes column. A colleague will review uncertain verbalizations later.
    \item If you have the necessity to check the meaning or the translation of certain entities of claim you can use WikiData \nolinkurl{https://www.wikidata.org/wiki/Wikidata:Main_Page}.
    \item Named entities
    \begin{itemize}
        \item For named entities, if an equivalent exists in Italian or Spanish, please use the correct translation in your verbalization.
        \item If you have doubts about the translation of a claim, you can find the corresponding version in Spanish or Italian on Wikidata.
        \item If you cannot find an equivalent in Italian or Spanish, do not translate it. Instead, flag the triple as problematic in the “Notes” column.
    \end{itemize}
\end{itemize}

\subsection{Validation}
\label{sec:validation}

In this phase, the annotators will cross-check and review colleagues’ annotations, evaluating whether each verbalization is acceptable by answering the following questions:
\begin{itemize}
    \item Does the text sound fluent and natural?
    \item Does the text include all and only the information from the data?
    \item Is the text written in good {language} (no spelling or grammatical mistakes)?
\end{itemize}

\section{Data Unit distribution per Category}
\label{sec:categories}

\begin{table}[h]
\centering
\begin{tabular}{l r}
\toprule
\textbf{Category} & \textbf{Count} \\
\midrule
Politician & 47 \\
Airport & 46 \\
Model & 30 \\
SportsTeam & 37 \\
Building & 47 \\
Company & 44 \\
Artist & 41 \\
City & 47 \\
MeanOfTransportation & 20 \\
Food & 35 \\
CelestialBody & 17 \\
Astronaut & 33 \\
Monument & 47 \\
ComicsCharacter & 48 \\
River & 30 \\
University & 47 \\
WrittenWork & 26 \\
Athlete & 47 \\
\midrule
\textbf{Type} & \textbf{Count} \\
\midrule
Long-tail & 408 \\
Head & 218 \\
\midrule
\textbf{Total entries} & \textbf{626} \\
\bottomrule
\end{tabular}
\caption{Distribution of data entries across categories and entity types.}
\label{tab:category_stats}
\end{table}

\section{Gold and Silver Entities Distribution}
\label{sec:Gold silver distribution}

\begin{table}[ht]
    \small
    \centering
    \begin{tabular}{lrrr}
        \toprule
        \textbf{Language} & \textbf{Gold} & \textbf{Silver} & \textbf{Total} \\
        \midrule
        English & $168$ & $447$ & $615$ \\
        Spanish & $173$ & $442$ & $615$ \\
        Italian & $384$ & $231$ & $615$ \\
        \midrule
        Total & $725$ & $1,120$ & $1,845$ \\
        \bottomrule
    \end{tabular}
    \caption{Distribution of gold and silver verbalizations by language in TailNLG.}
    \label{tab:verb_distribution}
\end{table} 

\section{Translation}

\subsection{Error correction}
\label{sec:error correction}

\begin{quote}
    You are provided with a Source, Translation, Translation quality analysis, and Translation quality score (weak, moderate, good, excellent). The Translation quality analysis contain a translation with marked error spans with different levels of severity (minor, major or critical). Additionally, we may provide a **reference translation**. Given this information, generate an explanation for each error and a fully correct translation.\\
    \textbf{Italian source}: Osvaldo il congilio è un personaggio dell'opera Lupo Alberto\\
    \textbf{English translation}: Osvaldo the congil is a character from the opera Wolf Albert.\\
    \textbf{Translation quality analysis}: Osvaldo the con<error severity="minor">gil</error> is a character from the opera<error severity="minor">Wolf Albert</error>.\\
    \textbf{Translation quality score}: good
\end{quote}

The corresponding output is:

\begin{quote}
    Explanation for error: The term "congilio" is a typographical error and should be corrected to "coniglio," which means "rabbit" in Italian. The correct translation of "Lupo Alberto" is "Wolfy Albert," as "Lupo" means "wolf" and "Alberto" is a common name in Italian. The translation also misses the gender of "Osvaldo," which is masculine, and the correct translation should reflect that. Additionally, the translation of "opera" as "opera" is correct, but the translation of "Wolf Albert" as "Wolf Albert" is incorrect because it does not convey the correct name of the work. The correct translation should reflect the proper name of the work, which is "Lupo Alberto."
\end{quote}

\subsection{Guidelines Human MT Evaluation}
\label{sec:MT assessment}
Annotators were instructed to strictly follow the evaluation guidelines and, when necessary, to correct the MT output by providing a  correct or more accurate version. 

Annotate as \textbf{1} if the error is present and \textbf{0} if not, respecting the following taxonomy: 

\subsection*{Accuracy}
Issues related to whether the information content of the target is equivalent to the source. Subcategories:  
\begin{itemize}
    \item \textbf{Terminology:} Issues related to the use of domain-specific terms.
    \item \textbf{Mistranslation:} Issues related to the improper translation of content, including erroneous translation of named entities.
    \item \textbf{Omission:} Content present in the source is missing in the target.
    \item \textbf{Addition:} Content not present in the source has been added to the target.
    \item \textbf{Untranslated:} Text inappropriately appears in the source language.
\end{itemize}

\subsection*{Fluency}
Issues related to the linguistic properties of the target without relation to its status as a translation. Subcategories:  
\begin{itemize}
    \item \textbf{Grammar:} Issues related to grammatical properties of the text.
    \item \textbf{Style:} The text shows stylistic problems.
    \item \textbf{Spelling:} The text is spelled incorrectly.
    \item \textbf{Typography:} Problems related to typographical conventions.
    \item \textbf{Unintelligible:} Text is garbled or otherwise unintelligible; indicates a major breakdown in fluency.
\end{itemize}

\subsection*{Corrected Translation}
If the target translation is incorrect, you may propose a new translation in the ``Corrected translation'' column.

\textbf{Guidelines for Annotators}
\begin{itemize}
    \item Evaluate whether translating a Named Entity (NE) makes sense in context, taking into account cultural and linguistic conventions; some NEs may be left untranslated.
    \item Named entities should be considered correctly translated when it is common in the target language to refer to the entity in that way; otherwise, mark them as incorrect. Examples:
    \begin{itemize}
        \item Crist Redemptor $\rightarrow$ Cristo Redentore $\rightarrow$ Christ the Redeemer [Correct]
        \item Lupo Alberto $\rightarrow$ Wolfy Albert [Incorrect]
        \item Louis the Springer $\rightarrow$ Luigi il Precursore [Incorrect]
    \end{itemize}
    Wrong NEs must be annotated in the column ``Mistranslation''.
    \item Assess when it is appropriate to keep a term in English (or another source language).
    \item Pay attention to the correct use of verb tenses.
    \item Ensure agreement between singular and plural.
    \item The translation should be read naturally in the target language.
    \item Ensure that the meaning is accurately preserved.
    \item Annotators should mark \textbf{1} for each subcategory when an issue occurs. If you label \textbf{1}, specify why in the ``note'' column.
\end{itemize}

\subsection{Manual assessment of translation}
\label{sec:Manual assessment of translation}
In the table below are reported the Spearman and Pearson correlation between the two annotators.

\begin{table}[]
    \centering
    \small
    \begin{tabular}{cc|cc}
        \toprule
        \textbf{Source} & \textbf{Target} & $\rho$ & $r$ \\
        \midrule
        
        es & en & 0.23 & 0.25 \\
        en & es & \,\,\,0.32* & 0.27\\
        it & en & 0.15 & 0.02\\
        en & it & \,\,\,0.56* & 0.64\\
        es & it & 0.26 & 0.22\\
        it & es & \,\,\,0.53* & \,\,\,0.49*\\
        
        \bottomrule
    \end{tabular}
    \caption{Spearman ($\rho$) and Pearson ($r$) correlation coefficients across language pairs. Source and Target denote the source language and the target language into which the source is translated, respectively. Asterisks (*) denote a statistically significant difference (p < 0.05).}
    \label{tab:iaa_correlation_new}
\end{table}

\section{TailNLG Benchmark}
\label{app:tailnlg}

Each entry contains the following metadata labels:
\begin{itemize}
  \item \textbf{Category}: domain/class of the subject (e.g., Artist, City, Monument).
  \item \textbf{Eid}: internal entry id (e.g., \texttt{Id1}).
  \item \textbf{Shape\_type}: data Units configuration (chain, sibling, mixed) that controls structural complexity.
  \item \textbf{Size}: number of triples for Data Units.
  \item \textbf{Qid}: Wikidata Id.
  \item \textbf{Type}: Distinction between head and long-tail entities
  \item \textbf{Sub\_type}: Distinction between subtypes (e.g. head entities with rare claims).
  \item \textbf{triplesets}: \texttt{originaltripleset} (as collected) and \texttt{modifiedtripleset} (normalized input).
  \item \textbf{lexicalizations}: one or more \texttt{lex} elements with attributes \texttt{quality} (gold/silver) and \texttt{lang} (en/es/it).
\end{itemize}

\section{Prompts}
\label{sec:prompts}

\paragraph{English version}
In English, structured data is commonly represented as triples, with the format
[subject, predicate, object]. Based on these triples, generate a single-paragraph text composed of complete, grammatically correct, and natural sentences. 
INSTRUCTIONS: \\
- Generate the text solely from the input triples \\
- Return the final verbalization with this format: The final verbalization is: [verbalization output] \\
- Insert the verbalization of the input triples within square brackets [], without adding anything else. 
\paragraph{Spanish version}
En español, los datos estructurados se representan comúnmente como tríos, con el formato [sujeto, predicado, objeto]. Basándose en estos tríos, genere un texto de un solo párrafo compuesto por oraciones completas, gramaticalmente correctas y naturales.\\
INSTRUCCIONES: \\
- Genere el texto únicamente a partir de las tripletas de entrada. \\
- Devuelva la verbalización final con este formato: La verbalización final es: [salida de verbalización] \\
- entre corchetes [] inserte la verbalización de las tripletas de entrada, sin añadir nada más 
\paragraph{Italian version}
In italiano, i dati strutturati sono comunemente rappresentati come triple, con il formato 
[soggetto, predicato, oggetto]. Sulla base di queste triple, genera un testo di un solo paragrafo composto da frasi complete, grammaticalmente corrette e naturali. \\
INSTRUZIONI: \\
- Genera il testo esclusivamente dalle triple in input \\
- Restituisci la verbalizzazione finale con questo formato: La verbalizzazione finale è: [output di verbalizzazione] \\
- tra le parentesi quadre [] inserisci la verbalizzazione delle triple in input, senza aggiungere altro.

\section{Results}
\label{sec:results appendix}

\begin{table*}[ht]
\footnotesize
\centering
\begin{tabular}{l l | c c c c c c c}
\toprule
\textbf{Model} & \textbf{Type} & \textbf{BERT$_{r}$} & \textbf{BLEU}~$\uparrow$ & \textbf{chrF}~$\uparrow$ & \textbf{R-1}~$\uparrow$ & \textbf{R-2}~$\uparrow$ & \textbf{R-L}~$\uparrow$ & \textbf{PPL}~$\downarrow$ \\
\midrule
\multirow{2}{*}{Qwen-3B} 
& Long-tail & \cellcolor[HTML]{F2F2F2}0.50 & \cellcolor[HTML]{F2F2F2}0.13 & \cellcolor[HTML]{F2F2F2}50.10 & \cellcolor[HTML]{F2F2F2}0.53 & \cellcolor[HTML]{F2F2F2}0.30 & \cellcolor[HTML]{F2F2F2}0.45 & \cellcolor[HTML]{F2F2F2}182.87 \\
 & Top-head  & 0.51 & 0.13 & 49.12 & 0.52 & 0.29 & 0.44 & \,\,\,70.45\textsuperscript{*} \\
\midrule
\multirow{2}{*}{Qwen-7B} 
 & Long-tail & \cellcolor[HTML]{F2F2F2}0.56 & \cellcolor[HTML]{F2F2F2}0.19 & \cellcolor[HTML]{F2F2F2}54.27 & \cellcolor[HTML]{F2F2F2}0.58 & \cellcolor[HTML]{F2F2F2}0.35 & \cellcolor[HTML]{F2F2F2}0.49 & \cellcolor[HTML]{F2F2F2}138.70 \\
 & Top-head  & 0.58 & 0.18 & 53.41 & 0.57 & 0.35 & 0.48 & \,\,\,87.96\textsuperscript{*} \\
\midrule
\multirow{2}{*}{Gemma-4b} 
 & Long-tail & \cellcolor[HTML]{F2F2F2}0.53 & \cellcolor[HTML]{F2F2F2}0.17 & \cellcolor[HTML]{F2F2F2}54.32 & \cellcolor[HTML]{F2F2F2}0.57 & \cellcolor[HTML]{F2F2F2}0.35 & \cellcolor[HTML]{F2F2F2}0.48 & \cellcolor[HTML]{F2F2F2}4724.94 \\
 & Top-head  & 0.56 & 0.17 & 53.48 & 0.56 & 0.34 & 0.46 & \,\,\,3488.00\textsuperscript{*} \\
\midrule
\multirow{2}{*}{Gemma-12b} 
 & Long-tail & \cellcolor[HTML]{F2F2F2}0.62 & \cellcolor[HTML]{F2F2F2}0.24 & \cellcolor[HTML]{F2F2F2}60.38 & \cellcolor[HTML]{F2F2F2}0.64 & \cellcolor[HTML]{F2F2F2}0.42 & \cellcolor[HTML]{F2F2F2}0.55 & \cellcolor[HTML]{F2F2F2}455.57\textsuperscript{*} \\
 & Top-head  & 0.64 & 0.23 & 58.73 & 0.63 & 0.40 & 0.53 & 797.47 \\
\midrule
\multirow{2}{*}{Llama-3B} 
 & Long-tail & \cellcolor[HTML]{F2F2F2}0.49 & \cellcolor[HTML]{F2F2F2}0.13 & \cellcolor[HTML]{F2F2F2}49.75 & \cellcolor[HTML]{F2F2F2}0.51 & \cellcolor[HTML]{F2F2F2}0.30 & \cellcolor[HTML]{F2F2F2}0.43 & \cellcolor[HTML]{F2F2F2}102.40 \\
 & Top-head  & 0.50 & 0.14 & 48.19 & 0.49 & 0.29 & 0.42 & \,\,\,72.55\textsuperscript{*} \\
\midrule
\multirow{2}{*}{Llama-8B} 
 & Long-tail & \cellcolor[HTML]{F2F2F2}0.55 & \cellcolor[HTML]{F2F2F2}0.18 & \cellcolor[HTML]{F2F2F2}53.81 & \cellcolor[HTML]{F2F2F2}0.58 & \cellcolor[HTML]{F2F2F2}0.36 & \cellcolor[HTML]{F2F2F2}0.50 & \cellcolor[HTML]{F2F2F2}134.53 \\
 & Top-head  & 0.57 & 0.19 & 53.34 & 0.57 & 0.36 & 0.48 & \,\,\,74.13\textsuperscript{*} \\
\bottomrule
\end{tabular}
\caption{Comparison of models by type. Asterisks ($^{*}$) denote a statistically significant difference with $p < 0.05$.}
\label{tab:type_comparison_full}
\end{table*}

\begin{table*}[]
\footnotesize
\centering
\begin{tabular}{l l | c c c c c c c c }
\toprule
\textbf{Model} & \textbf{Lang} & \textbf{BERT$_{r}$} & \textbf{BLEU}~$\uparrow$ & \textbf{chrF}~$\uparrow$ & \textbf{R-1}~$\uparrow$ & \textbf{R-2}~$\uparrow$ & \textbf{R-L}~$\uparrow$ & \textbf{PPL}~$\downarrow$ \\
\midrule
\multirow{3}{*}{Llama-3B} & en &  \,\,\cellcolor[HTML]{F2F2F2}0.58\textsuperscript{*} &  \,\,\cellcolor[HTML]{F2F2F2}0.21\textsuperscript{*} & \,\,\cellcolor[HTML]{F2F2F2}56.67\textsuperscript{*} &  \,\,\cellcolor[HTML]{F2F2F2}0.63\textsuperscript{*} &  \,\,\cellcolor[HTML]{F2F2F2}0.40\textsuperscript{*} & \,\,\cellcolor[HTML]{F2F2F2}0.53\textsuperscript{*} & \,\,\cellcolor[HTML]{F2F2F2}73.51\textsuperscript{*} \\
& es &  0.46 &  0.11 & 46.40 &  0.47 &  0.26 &  0.39 & 114.99 \\
& it &  \cellcolor[HTML]{F2F2F2}0.44 &  \cellcolor[HTML]{F2F2F2}0.08 & \cellcolor[HTML]{F2F2F2}44.26 &  \cellcolor[HTML]{F2F2F2}0.42 &  \cellcolor[HTML]{F2F2F2}0.22 &  \cellcolor[HTML]{F2F2F2}0.35 & \cellcolor[HTML]{F2F2F2}81.98 \\
\midrule
\multirow{3}{*}{Llama-8B} & en &\,\,\cellcolor[HTML]{F2F2F2}0.63\textsuperscript{*} &\,\,\cellcolor[HTML]{F2F2F2}0.25\textsuperscript{*} &\,\,\cellcolor[HTML]{F2F2F2}60.72\textsuperscript{*} &\,\,\cellcolor[HTML]{F2F2F2}0.66\textsuperscript{*} &\,\,\cellcolor[HTML]{F2F2F2}0.43\textsuperscript{*} &\,\,\cellcolor[HTML]{F2F2F2}0.56\textsuperscript{*} &\,\,\cellcolor[HTML]{F2F2F2}79.37\textsuperscript{*} \\
& es & 0.59 & 0.18 & 53.27 & 0.58 & 0.35 &  0.49 & 170.14 \\
& it &\cellcolor[HTML]{F2F2F2}0.45 &\cellcolor[HTML]{F2F2F2}0.13 &\cellcolor[HTML]{F2F2F2}46.85 &\cellcolor[HTML]{F2F2F2}0.48 &\cellcolor[HTML]{F2F2F2}0.28 &\cellcolor[HTML]{F2F2F2}0.42 &\cellcolor[HTML]{F2F2F2}80.23  \\
\midrule
\multirow{3}{*}{Qwen-3B} & en &  \cellcolor[HTML]{F2F2F2}0.50 &  \,\,\cellcolor[HTML]{F2F2F2}0.17\textsuperscript{*} & \,\,\cellcolor[HTML]{F2F2F2}52.79\textsuperscript{*} &  \,\,\cellcolor[HTML]{F2F2F2}0.60\textsuperscript{*} &  \,\,\cellcolor[HTML]{F2F2F2}0.36\textsuperscript{*} &  \,\,\cellcolor[HTML]{F2F2F2}0.51\textsuperscript{*} & \cellcolor[HTML]{F2F2F2}84.29 \\
& es &  \,\,0.54\textsuperscript{*} &  0.12 & 49.10 &  0.51 &  0.28 &  0.43 & 227.52\\
& it &  \cellcolor[HTML]{F2F2F2}0.48 &  \cellcolor[HTML]{F2F2F2}0.09 & \cellcolor[HTML]{F2F2F2}47.19 &  \cellcolor[HTML]{F2F2F2}0.46 &  \cellcolor[HTML]{F2F2F2}0.24 &  \cellcolor[HTML]{F2F2F2}0.39 & \cellcolor[HTML]{F2F2F2}100.23 \\
\midrule
\multirow{3}{*}{Qwen-7B} & en &\,\,\cellcolor[HTML]{F2F2F2}0.67\textsuperscript{*} &\,\,\cellcolor[HTML]{F2F2F2}0.29\textsuperscript{*} &\,\,\cellcolor[HTML]{F2F2F2}62.71\textsuperscript{*} &\,\,\cellcolor[HTML]{F2F2F2}0.69\textsuperscript{*} &\,\,\cellcolor[HTML]{F2F2F2}0.46\textsuperscript{*} &\,\,\cellcolor[HTML]{F2F2F2}0.59\textsuperscript{*}  &\cellcolor[HTML]{F2F2F2}119.43\\
& es &  0.60 &  0.16 & 52.78 &  0.56 &  0.33 &  0.48  & 149.16 \\
& it &\cellcolor[HTML]{F2F2F2}0.45 &\cellcolor[HTML]{F2F2F2}0.11 &\cellcolor[HTML]{F2F2F2}46.25 &\cellcolor[HTML]{F2F2F2}0.46 &\cellcolor[HTML]{F2F2F2}0.26 &\cellcolor[HTML]{F2F2F2}0.40  &\,\,\cellcolor[HTML]{F2F2F2}85.39\textsuperscript{*}\\
\midrule
\multirow{3}{*}{Gemma-4b} & en &  \,\,\cellcolor[HTML]{F2F2F2}0.62\textsuperscript{*} &  \,\,\cellcolor[HTML]{F2F2F2}0.23\textsuperscript{*} & \,\,\cellcolor[HTML]{F2F2F2}60.32\textsuperscript{*} &  \,\,\cellcolor[HTML]{F2F2F2}0.66\textsuperscript{*} &  \,\,\cellcolor[HTML]{F2F2F2}0.42\textsuperscript{*} &  \,\,\cellcolor[HTML]{F2F2F2}0.55\textsuperscript{*} & \cellcolor[HTML]{F2F2F2}3798.99 \\
& es &  0.51 &  0.14 & 50.76 &  0.53 &  0.31 &  0.44 & 6068.41 \\
& it &  \cellcolor[HTML]{F2F2F2}0.50 &  \cellcolor[HTML]{F2F2F2}0.14 & \cellcolor[HTML]{F2F2F2}50.85 &  \cellcolor[HTML]{F2F2F2}0.50 &  \cellcolor[HTML]{F2F2F2}0.30 &  \cellcolor[HTML]{F2F2F2}0.43 & \,\,\cellcolor[HTML]{F2F2F2}2794.93\textsuperscript{*} \\
\midrule
\multirow{3}{*}{Gemma-12b} & en &\,\,\cellcolor[HTML]{F2F2F2}0.64\textsuperscript{*} &\,\,\cellcolor[HTML]{F2F2F2}0.27\textsuperscript{*} &\,\,\cellcolor[HTML]{F2F2F2}61.68\textsuperscript{*} &\,\,\cellcolor[HTML]{F2F2F2}0.68\textsuperscript{*} &\,\,\cellcolor[HTML]{F2F2F2}0.44\textsuperscript{*} &\,\,\cellcolor[HTML]{F2F2F2}0.57\textsuperscript{*} &\cellcolor[HTML]{F2F2F2}357.51 \\
& es &  0.63 &  0.21 & 58.27 &  0.62 &  0.39 &  0.52 & 739.90\\
& it &\cellcolor[HTML]{F2F2F2}0.61 &\cellcolor[HTML]{F2F2F2}0.22 &\cellcolor[HTML]{F2F2F2}59.17 &\cellcolor[HTML]{F2F2F2}0.62 &\cellcolor[HTML]{F2F2F2}0.40 &\cellcolor[HTML]{F2F2F2}0.53  &\cellcolor[HTML]{F2F2F2}687.37\\
\bottomrule
\end{tabular}
\caption{Average scores for different languages for each model (zero-shot).  Asterisks ($^{*}$) denote a statistically significant difference with $p < 0.05$.}
\label{tab:results_per_language_full}
\end{table*}

\begin{table*}[]
\small
\centering
\begin{tabular}{l l | c c c c c c c c}
\toprule
\textbf{Model} & \textbf{Quality} & \textbf{BERT$_{r}$}~$\uparrow$ & \textbf{BLEU}~$\uparrow$ & \textbf{chrF}~$\uparrow$ & \textbf{R-1}~$\uparrow$ & \textbf{R-2}~$\uparrow$ & \textbf{R-L}~$\uparrow$ & \textbf{PPL}~$\downarrow$ \\
\midrule
\multirow{2}{*}{Llama-3B} & gold & \cellcolor[HTML]{F2F2F2}0.46 & \cellcolor[HTML]{F2F2F2}0.12 & \cellcolor[HTML]{F2F2F2}48.04 & \cellcolor[HTML]{F2F2F2}0.48 & \cellcolor[HTML]{F2F2F2}0.27 & \cellcolor[HTML]{F2F2F2}0.40 & \cellcolor[HTML]{F2F2F2}95.30 \\
& silver  & 0.52\textsuperscript{*} & 0.14 & 49.80 & 0.52\textsuperscript{*} & 0.30\textsuperscript{*} & 0.44\textsuperscript{*} & 86.86\textsuperscript{*} \\

\midrule

\multirow{2}{*}{Llama-8B} & gold & \cellcolor[HTML]{F2F2F2}0.52 & \cellcolor[HTML]{F2F2F2}0.18 & \cellcolor[HTML]{F2F2F2}52.68 & \cellcolor[HTML]{F2F2F2}0.50 & \cellcolor[HTML]{F2F2F2}0.55\textsuperscript{*} & \cellcolor[HTML]{F2F2F2}0.48 & \cellcolor[HTML]{F2F2F2}90.94 \\
& silver  & 0.58\textsuperscript{*} & 0.19 & 54.22 & 0.59\textsuperscript{*} & 0.36 & 0.50\textsuperscript{*} & 122.10\textsuperscript{*} \\

\midrule
\multirow{2}{*}{Qwen-3B} & gold  & \cellcolor[HTML]{F2F2F2}0.50 & \cellcolor[HTML]{F2F2F2}0.13 & \cellcolor[HTML]{F2F2F2}50.62 & \cellcolor[HTML]{F2F2F2}0.53 & \cellcolor[HTML]{F2F2F2}0.30 & \cellcolor[HTML]{F2F2F2}0.44 & \cellcolor[HTML]{F2F2F2}94.12 \\
& silver  & 0.51 & 0.13 & 49.08 & 0.52 & 0.29 & 0.45 & 165.11 \\

\midrule
\multirow{2}{*}{Qwen-7B} & gold  & \cellcolor[HTML]{F2F2F2}0.53 & \cellcolor[HTML]{F2F2F2}0.17 & \cellcolor[HTML]{F2F2F2}52.59 & \cellcolor[HTML]{F2F2F2}0.55 & \cellcolor[HTML]{F2F2F2}0.33 & \cellcolor[HTML]{F2F2F2}0.47 & \cellcolor[HTML]{F2F2F2}117.53 \\
& silver  & 0.60\textsuperscript{*} & 0.20 & 54.75 & 0.59\textsuperscript{*} & 0.36 & 0.50 & 118.35 \\

\midrule
\multirow{2}{*}{Gemma-4b} & gold  & \cellcolor[HTML]{F2F2F2}0.52 & \cellcolor[HTML]{F2F2F2}0.17 & \cellcolor[HTML]{F2F2F2}53.67 & \cellcolor[HTML]{F2F2F2}0.55 & \cellcolor[HTML]{F2F2F2}0.34 & \cellcolor[HTML]{F2F2F2}0.46 & \cellcolor[HTML]{F2F2F2}1143.13 \\
& silver  & 0.55\textsuperscript{*} & 0.18 & 54.18 & 0.57 & 0.35 & 0.48 & 6197.47 \\

\midrule
\multirow{2}{*}{Gemma-12b} & gold  & \cellcolor[HTML]{F2F2F2}0.61 & \cellcolor[HTML]{F2F2F2}0.23 & \cellcolor[HTML]{F2F2F2}59.88 & \cellcolor[HTML]{F2F2F2}0.63 & \cellcolor[HTML]{F2F2F2}0.41 & \cellcolor[HTML]{F2F2F2}0.54 & \cellcolor[HTML]{F2F2F2}406.83 \\
& silver  & 0.64\textsuperscript{*} & 0.24 & 59.59 & 0.64 & 0.41 & 0.54 & 715.74 \\

\bottomrule
\end{tabular}
\caption{Average scores for different quality levels (silver, gold) for each model (zero-shot). Asterisks ($^{*}$) denote a statistically significant difference with $p < 0.05$.}
\label{tab:results_per_quality}
\end{table*}

\begin{table*}[]
\centering
\footnotesize
\begin{tabular}{ll  ccccccc}

\toprule
\textbf{Type} & \textbf{Quality} & \textbf{BERT$_{r}$} & \textbf{BLEU}~$\uparrow$ & \textbf{chrF}~$\uparrow$ & \textbf{R-1}~$\uparrow$ & \textbf{R-2}~$\uparrow$ & \textbf{R-L}~$\uparrow$ & \textbf{PPL}~$\downarrow$ \\

\midrule
\midrule

\multicolumn{9}{c}{\textbf{Llama-3B}} \\

\midrule

\multirow{2}{*}{Long-tail} & \cellcolor[HTML]{F2F2F2}Silver & \cellcolor[HTML]{F2F2F2}\,\,0.50\textsuperscript{*} & \cellcolor[HTML]{F2F2F2}0.14 & \cellcolor[HTML]{F2F2F2}49.63 & \cellcolor[HTML]{F2F2F2}0.52 & \cellcolor[HTML]{F2F2F2}0.30 & \cellcolor[HTML]{F2F2F2}0.44 & \cellcolor[HTML]{F2F2F2}\,\,89.70\textsuperscript{*} \\
 & Gold & 0.47 & 0.13 & 49.95 & 0.50 & 0.29 & 0.42 & 122.69 \\

\midrule

\multirow{2}{*}{Top-head} & \cellcolor[HTML]{F2F2F2}Silver & \cellcolor[HTML]{F2F2F2}\,\,0.53\textsuperscript{*} & \cellcolor[HTML]{F2F2F2}\,\,0.16\textsuperscript{*} & \cellcolor[HTML]{F2F2F2}\,\,50.04\textsuperscript{*} & \cellcolor[HTML]{F2F2F2}\,\,0.52\textsuperscript{*} & \cellcolor[HTML]{F2F2F2}\,\,0.31\textsuperscript{*} & \cellcolor[HTML]{F2F2F2}\,\,0.44\textsuperscript{*} & \cellcolor[HTML]{F2F2F2}82.59 \\
 & Gold & 0.46 & 0.11 & 45.40 & 0.45 & 0.25 & 0.38 & 57.57 \\

\midrule
\midrule

\multicolumn{9}{c}{\textbf{Llama-8B}} \\

\midrule

\multirow{2}{*}{Long-tail} & \cellcolor[HTML]{F2F2F2}Silver & \cellcolor[HTML]{F2F2F2}0.56 & \cellcolor[HTML]{F2F2F2}0.18 & \cellcolor[HTML]{F2F2F2}53.64 & \cellcolor[HTML]{F2F2F2}0.58 & \cellcolor[HTML]{F2F2F2}0.35 & \cellcolor[HTML]{F2F2F2}0.50 & \cellcolor[HTML]{F2F2F2}149.01 \\
 & Gold & 0.52 & 0.19 & 54.07 & 0.57 & 0.36 & 0.49 & \,\,111.15\textsuperscript{*} \\

\midrule

\multirow{2}{*}{Top-head} & \cellcolor[HTML]{F2F2F2}Silver & \cellcolor[HTML]{F2F2F2}\,\,0.60\textsuperscript{*} & \cellcolor[HTML]{F2F2F2}\,\,0.21\textsuperscript{*} & \cellcolor[HTML]{F2F2F2}\,\,55.08\textsuperscript{*} & \cellcolor[HTML]{F2F2F2}\,\,0.60\textsuperscript{*} & \cellcolor[HTML]{F2F2F2}\,\,0.38\textsuperscript{*} & \cellcolor[HTML]{F2F2F2}\,\,0.50\textsuperscript{*} & \cellcolor[HTML]{F2F2F2}81.60 \\
 & Gold & 0.52 & 0.16 & 50.76 & 0.53 & 0.32 & 0.45 & 63.08 \\

\midrule
\midrule

\multicolumn{9}{c}{\textbf{Gemma-4B}} \\

\midrule

\multirow{2}{*}{Long-tail} & \cellcolor[HTML]{F2F2F2}Silver & \cellcolor[HTML]{F2F2F2}0.54 & \cellcolor[HTML]{F2F2F2}0.17 & \cellcolor[HTML]{F2F2F2}53.96 & \cellcolor[HTML]{F2F2F2}0.57 & \cellcolor[HTML]{F2F2F2}0.35 & \cellcolor[HTML]{F2F2F2}0.48 & \cellcolor[HTML]{F2F2F2}6781.26 \\
 & Gold & 0.52 & 0.18 & 54.91 & 0.56 & 0.35 & 0.48 & \,\,1405.88\textsuperscript{*} \\

\midrule

\multirow{2}{*}{Top-head} & \cellcolor[HTML]{F2F2F2}Silver & \cellcolor[HTML]{F2F2F2}\,\,0.58\textsuperscript{*} & \cellcolor[HTML]{F2F2F2}\,\,0.18\textsuperscript{*} & \cellcolor[HTML]{F2F2F2}\,\,54.50\textsuperscript{*} & \cellcolor[HTML]{F2F2F2}\,\,0.57\textsuperscript{*} & \cellcolor[HTML]{F2F2F2}\,\,0.35\textsuperscript{*} & \cellcolor[HTML]{F2F2F2}\,\,0.48\textsuperscript{*} & \cellcolor[HTML]{F2F2F2}5319.16 \\
 & Gold & 0.53 & 0.16 & 51.96 & 0.53 & 0.32 & 0.44 & 781.08 \\

\midrule
\midrule

\multicolumn{9}{c}{\textbf{Gemma-12B}} \\

\midrule

\multirow{2}{*}{Long-tail} & \cellcolor[HTML]{F2F2F2}Silver & \cellcolor[HTML]{F2F2F2}\,\,0.63\textsuperscript{*} & \cellcolor[HTML]{F2F2F2}0.24 & \cellcolor[HTML]{F2F2F2}59.82 & \cellcolor[HTML]{F2F2F2}0.64 & \cellcolor[HTML]{F2F2F2}0.41 & \cellcolor[HTML]{F2F2F2}0.55 & \cellcolor[HTML]{F2F2F2}460.31 \\
 & Gold & 0.61 & 0.24 & \,\,61.28\textsuperscript{*} & 0.65 & 0.43 & 0.56 & 447.92 \\

\midrule

\multirow{2}{*}{Top-head} & \cellcolor[HTML]{F2F2F2}Silver & \cellcolor[HTML]{F2F2F2}\,\,0.65\textsuperscript{*} & \cellcolor[HTML]{F2F2F2}0.24 & \cellcolor[HTML]{F2F2F2}59.26 & \cellcolor[HTML]{F2F2F2}\,\,0.64\textsuperscript{*} & \cellcolor[HTML]{F2F2F2}\,\,0.41\textsuperscript{*} & \cellcolor[HTML]{F2F2F2}\,\,0.54\textsuperscript{*} & \cellcolor[HTML]{F2F2F2}1100.04 \\
 & Gold & 0.61 & 0.21 & 57.95 & 0.61 & 0.39 & 0.51 & 350.20 \\

\midrule
\midrule

\multicolumn{9}{c}{\textbf{Qwen-3B}} \\

\midrule

\multirow{2}{*}{Long-tail} & \cellcolor[HTML]{F2F2F2}Silver & \cellcolor[HTML]{F2F2F2}0.50 & \cellcolor[HTML]{F2F2F2}0.12 & \cellcolor[HTML]{F2F2F2}48.93 & \cellcolor[HTML]{F2F2F2}0.52 & \cellcolor[HTML]{F2F2F2}0.29 & \cellcolor[HTML]{F2F2F2}0.44 & \cellcolor[HTML]{F2F2F2}224.95 \\
 & Gold & 0.50 & 0.15 & \,\,51.91\textsuperscript{*} & 0.54 & 0.31 & 0.46 & \,\,116.44\textsuperscript{*} \\

\midrule

\multirow{2}{*}{Top-head} & \cellcolor[HTML]{F2F2F2}Silver & \cellcolor[HTML]{F2F2F2}\,\,0.52\textsuperscript{*} & \cellcolor[HTML]{F2F2F2}0.13 & \cellcolor[HTML]{F2F2F2}49.29 & \cellcolor[HTML]{F2F2F2}\,\,0.53\textsuperscript{*} & \cellcolor[HTML]{F2F2F2}\,\,0.30\textsuperscript{*} & \cellcolor[HTML]{F2F2F2}\,\,0.45\textsuperscript{*} & \cellcolor[HTML]{F2F2F2}75.08 \\
 & Gold & 0.51 & 0.12 & 48.83 & 0.50 & 0.28 & 0.42 & 63.37 \\

\midrule
\midrule

\multicolumn{9}{c}{\textbf{Qwen-7B}} \\

\midrule

\multirow{2}{*}{Long-tail} & \cellcolor[HTML]{F2F2F2}Silver & \cellcolor[HTML]{F2F2F2}\,\,0.59\textsuperscript{*} & \cellcolor[HTML]{F2F2F2}0.19 & \cellcolor[HTML]{F2F2F2}54.66 & \cellcolor[HTML]{F2F2F2}0.59 & \cellcolor[HTML]{F2F2F2}0.36 & \cellcolor[HTML]{F2F2F2}0.50 & \cellcolor[HTML]{F2F2F2}\,\,129.26\textsuperscript{*} \\
 & Gold & 0.52 & 0.18 & 53.57 & 0.56 & 0.35 & 0.49 & 153.98 \\

\midrule

\multirow{2}{*}{Top-head} & \cellcolor[HTML]{F2F2F2}Silver & \cellcolor[HTML]{F2F2F2}\,\,0.62\textsuperscript{*} & \cellcolor[HTML]{F2F2F2}\,\,0.20\textsuperscript{*} & \cellcolor[HTML]{F2F2F2}\,\,54.89\textsuperscript{*} & \cellcolor[HTML]{F2F2F2}\,\,0.59\textsuperscript{*} & \cellcolor[HTML]{F2F2F2}\,\,0.36\textsuperscript{*} & \cellcolor[HTML]{F2F2F2}\,\,0.50\textsuperscript{*} & \cellcolor[HTML]{F2F2F2}101.93 \\
 & Gold & 0.54 & 0.16 & 51.23 & 0.53 & 0.32 & 0.45 & 67.31 \\

\bottomrule
\end{tabular}
\caption{Performance comparison of models broken down by entity type (long-tail / top-head) and annotation quality (silver / gold). Asterisks ($^{*}$) denote a statistically significant difference between silver and gold within the same type ($p < 0.05$, Mann-Whitney U test).}
\label{tab:multi_model_results}
\end{table*}

\begin{table*}[]
\centering
\footnotesize
\begin{tabular}{ll  ccccccc}

\toprule
\textbf{Type} & \textbf{Lang} & \textbf{BERT$_{r}$} & \textbf{BLEU}~$\uparrow$ & \textbf{chrF}~$\uparrow$ & \textbf{R-1}~$\uparrow$ & \textbf{R-2}~$\uparrow$ & \textbf{R-L}~$\uparrow$ & \textbf{PPL}~$\downarrow$ \\

\midrule
\midrule

\multicolumn{9}{c}{\textbf{Llama-3B}} \\

\midrule

\multirow{4}{*}{Long-tail} & en & \cellcolor[HTML]{F2F2F2}0.57 &\cellcolor[HTML]{F2F2F2}0.21 &\cellcolor[HTML]{F2F2F2}56.58 &\cellcolor[HTML]{F2F2F2}0.62 &\cellcolor[HTML]{F2F2F2}0.39 &\cellcolor[HTML]{F2F2F2}0.53 &\cellcolor[HTML]{F2F2F2}86.17 \\
& es & 0.45 & 0.11 & 47.07 & 0.48 & 0.26 & 0.40 & 144.99 \\
& it & \cellcolor[HTML]{F2F2F2}0.44 &\cellcolor[HTML]{F2F2F2}0.09 &\cellcolor[HTML]{F2F2F2}45.61 &\cellcolor[HTML]{F2F2F2}0.44 &\cellcolor[HTML]{F2F2F2}0.23 &\cellcolor[HTML]{F2F2F2}0.37 &\cellcolor[HTML]{F2F2F2}75.80 \\
& total & 0.49 & 0.13 & 49.75 & 0.51 & 0.30 & 0.43 & 102.32 \\

\midrule

\multirow{4}{*}{WebNLG} & en & \cellcolor[HTML]{F2F2F2}0.56 &\cellcolor[HTML]{F2F2F2}0.23 &\cellcolor[HTML]{F2F2F2}57.17 &\,\,\cellcolor[HTML]{F2F2F2}0.71\textsuperscript{*} &\,\,\cellcolor[HTML]{F2F2F2}0.47\textsuperscript{*} &\cellcolor[HTML]{F2F2F2}0.56 &\,\,\cellcolor[HTML]{F2F2F2}53.67\textsuperscript{*} \\
& es & \,\,\underline{0.49}\textsuperscript{*} & \,\,0.16\textsuperscript{*} & 49.20 & \,\,0.58\textsuperscript{*} & \,\,0.35\textsuperscript{*} & \,\,0.45\textsuperscript{*} & \,\,68.28\textsuperscript{*} \\
& it & \,\,\cellcolor[HTML]{F2F2F2}\underline{0.48}\textsuperscript{*} &\,\,\cellcolor[HTML]{F2F2F2}0.13\textsuperscript{*} &\,\,\cellcolor[HTML]{F2F2F2}47.75\textsuperscript{*} &\,\,\cellcolor[HTML]{F2F2F2}0.53\textsuperscript{*} &\,\,\cellcolor[HTML]{F2F2F2}0.30\textsuperscript{*} &\,\,\cellcolor[HTML]{F2F2F2}0.42\textsuperscript{*} &\,\,\cellcolor[HTML]{F2F2F2}41.18\textsuperscript{*} \\
& total & 0.51 & \,\,0.17\textsuperscript{*} & 51.37 & 0.60 & 0.37 & \,\,0.48\textsuperscript{*} & \,\,54.38\textsuperscript{*} \\

\midrule
\midrule

\multicolumn{9}{c}{\textbf{Llama-8B}} \\
\toprule

\multirow{4}{*}{Long-tail} & en & \cellcolor[HTML]{F2F2F2}0.61 &\cellcolor[HTML]{F2F2F2}0.25 &\cellcolor[HTML]{F2F2F2}60.39 &\cellcolor[HTML]{F2F2F2}0.66 &\cellcolor[HTML]{F2F2F2}0.43 &\cellcolor[HTML]{F2F2F2}0.56 &\cellcolor[HTML]{F2F2F2}93.36 \\
& es & \,\,0.58\textsuperscript{*} & 0.17 & 53.30 & 0.58 & 0.35 & 0.49 & 222.16 \\
& it & \,\,\cellcolor[HTML]{F2F2F2}0.46\textsuperscript{*} &\cellcolor[HTML]{F2F2F2}0.13 &\cellcolor[HTML]{F2F2F2}47.73 &\cellcolor[HTML]{F2F2F2}0.49 &\cellcolor[HTML]{F2F2F2}0.29 &\cellcolor[HTML]{F2F2F2}0.43 &\cellcolor[HTML]{F2F2F2}88.06 \\
& total & 0.55 & 0.18 & 53.81 & 0.58 & 0.36 & 0.50 & 134.53 \\

\midrule

\multirow{4}{*}{WebNLG} & en & \cellcolor[HTML]{F2F2F2}0.61 &\cellcolor[HTML]{F2F2F2}0.30 &\cellcolor[HTML]{F2F2F2}65.53 &\,\,\cellcolor[HTML]{F2F2F2}0.74\textsuperscript{*} &\,\,\cellcolor[HTML]{F2F2F2}0.51\textsuperscript{*} &\,\,\cellcolor[HTML]{F2F2F2}0.59\textsuperscript{*} &\,\,\cellcolor[HTML]{F2F2F2}42.56\textsuperscript{*} \\
& es & 0.55 & \,\,0.19\textsuperscript{*} & 52.43 & \,\,0.66\textsuperscript{*} & \,\,0.43\textsuperscript{*} & \,\,0.53\textsuperscript{*} & \,\,90.42\textsuperscript{*} \\
& it & \cellcolor[HTML]{F2F2F2}0.43 &\,\,\cellcolor[HTML]{F2F2F2}0.15\textsuperscript{*} &\cellcolor[HTML]{F2F2F2}45.35 &\,\,\cellcolor[HTML]{F2F2F2}0.54\textsuperscript{*} &\,\,\cellcolor[HTML]{F2F2F2}0.34\textsuperscript{*} &\,\,\cellcolor[HTML]{F2F2F2}0.46\textsuperscript{*} &\,\,\cellcolor[HTML]{F2F2F2}71.73\textsuperscript{*} \\
& total & 0.53 & \,\,0.21\textsuperscript{*} & 53.77 & \,\,0.65\textsuperscript{*} & \,\,0.43\textsuperscript{*} &  \,\,0.53\textsuperscript{*} & \,\,68.23\textsuperscript{*} \\

\midrule
\midrule

\multicolumn{9}{c}{\textbf{Gemma-4B}} \\
\toprule

\multirow{4}{*}{Long-tail} & en & \cellcolor[HTML]{F2F2F2}0.60 &\cellcolor[HTML]{F2F2F2}0.23 &\cellcolor[HTML]{F2F2F2}59.85 &\cellcolor[HTML]{F2F2F2}0.66 &\cellcolor[HTML]{F2F2F2}0.42 &\cellcolor[HTML]{F2F2F2}0.55 &\cellcolor[HTML]{F2F2F2}4886.76 \\
& es & 0.50 & 0.14 & 51.49 & 0.54 & 0.32 & 0.45 & 8074.87 \\
& it & \cellcolor[HTML]{F2F2F2}0.49 &\cellcolor[HTML]{F2F2F2}0.15 &\cellcolor[HTML]{F2F2F2}51.62 &\cellcolor[HTML]{F2F2F2}0.51 &\cellcolor[HTML]{F2F2F2}0.31 &\cellcolor[HTML]{F2F2F2}0.44 &\cellcolor[HTML]{F2F2F2}1213.20 \\
& total & 0.53 & 0.17 & 54.32 & 0.57 & 0.35 & 0.48 & 4724.94 \\

\midrule

\multirow{4}{*}{WebNLG} & en & \cellcolor[HTML]{F2F2F2}0.61 &\,\,\cellcolor[HTML]{F2F2F2}0.28\textsuperscript{*} &\,\,\cellcolor[HTML]{F2F2F2}62.43\textsuperscript{*} &\,\,\cellcolor[HTML]{F2F2F2}0.74\textsuperscript{*} &\,\,\cellcolor[HTML]{F2F2F2}0.50\textsuperscript{*} &\,\,\cellcolor[HTML]{F2F2F2}0.59\textsuperscript{*} &\,\,\cellcolor[HTML]{F2F2F2}930.34\textsuperscript{*} \\
& es & 0.52 & \,\,0.20\textsuperscript{*} & \,\,53.79\textsuperscript{*} & \,\,0.61\textsuperscript{*} & \,\,0.39\textsuperscript{*} & \,\,0.48\textsuperscript{*} & \,\,1493.20\textsuperscript{*} \\
& it & \cellcolor[HTML]{F2F2F2}0.49 &\cellcolor[HTML]{F2F2F2}\,\,0.18\textsuperscript{*} &\cellcolor[HTML]{F2F2F2}53.72 &\,\,\cellcolor[HTML]{F2F2F2}0.57\textsuperscript{*} &\,\,\cellcolor[HTML]{F2F2F2}0.36\textsuperscript{*} &\cellcolor[HTML]{F2F2F2}0.46 &\,\,\cellcolor[HTML]{F2F2F2}601.97\textsuperscript{*} \\
& total & 0.54 & \,\,0.22\textsuperscript{*} & \,\,56.65\textsuperscript{*} & \,\,0.64\textsuperscript{*} & \,\,0.42\textsuperscript{*} & 0.51 & \,\,1008.50\textsuperscript{*} \\

\midrule
\midrule

\multicolumn{9}{c}{\textbf{Gemma-12B}} \\
\toprule
\multirow{4}{*}{Long-tail} & en & \cellcolor[HTML]{F2F2F2}0.63 &\cellcolor[HTML]{F2F2F2}0.26 &\cellcolor[HTML]{F2F2F2}61.65 &\cellcolor[HTML]{F2F2F2}0.67 &\cellcolor[HTML]{F2F2F2}0.44 &\cellcolor[HTML]{F2F2F2}0.58 &\,\,\cellcolor[HTML]{F2F2F2}330.02\textsuperscript{*} \\
& es & 0.62 & 0.21 & 58.80 & 0.62 & 0.40 & 0.53 & 611.42 \\
& it & \cellcolor[HTML]{F2F2F2}0.61 &\cellcolor[HTML]{F2F2F2}0.24 &\cellcolor[HTML]{F2F2F2}60.68 &\cellcolor[HTML]{F2F2F2}0.64 &\cellcolor[HTML]{F2F2F2}0.42 &\cellcolor[HTML]{F2F2F2}0.55 &\cellcolor[HTML]{F2F2F2}425.27 \\
& total & 0.62 & 0.24 & 60.38 & 0.64 & 0.42 & 0.55 & 455.57 \\

\midrule
\multirow{4}{*}{WebNLG} & en & \,\,\cellcolor[HTML]{F2F2F2}0.66\textsuperscript{*} &\,\,\cellcolor[HTML]{F2F2F2}0.33\textsuperscript{*} &\cellcolor[HTML]{F2F2F2}65.97 &\,\,\cellcolor[HTML]{F2F2F2}0.77\textsuperscript{*} &\,\,\cellcolor[HTML]{F2F2F2}0.53\textsuperscript{*} &\,\,\cellcolor[HTML]{F2F2F2}0.62\textsuperscript{*} &\cellcolor[HTML]{F2F2F2}333.93 \\
& es & 0.65 & \,\,0.30\textsuperscript{*} & 62.68 & \,\,0.71\textsuperscript{*} & \,\,0.49\textsuperscript{*} & \,\,0.57\textsuperscript{*} & \,\,210.04\textsuperscript{*} \\
& it & \,\,\cellcolor[HTML]{F2F2F2}0.64\textsuperscript{*} &\,\,\cellcolor[HTML]{F2F2F2}0.28\textsuperscript{*} &\,\,\cellcolor[HTML]{F2F2F2}62.24\textsuperscript{*} &\,\,\cellcolor[HTML]{F2F2F2}0.70\textsuperscript{*} &\,\,\cellcolor[HTML]{F2F2F2}0.47\textsuperscript{*} &\cellcolor[HTML]{F2F2F2}0.57 &\,\,\cellcolor[HTML]{F2F2F2}326.33\textsuperscript{*} \\
& total & \,\,0.65\textsuperscript{*} & \,\,0.30\textsuperscript{*} & 63.63 & \,\,0.73\textsuperscript{*} & \,\,0.50\textsuperscript{*} & \,\,0.59\textsuperscript{*} & \,\,290.10\textsuperscript{*} \\

\midrule
\midrule

\multicolumn{9}{c}{\textbf{Qwen-3B}} \\
\toprule

\multirow{4}{*}{Long-tail} & en & \cellcolor[HTML]{F2F2F2}0.49 &\cellcolor[HTML]{F2F2F2}0.17 &\cellcolor[HTML]{F2F2F2}52.66 &\cellcolor[HTML]{F2F2F2}0.59 &\cellcolor[HTML]{F2F2F2}0.35 &\cellcolor[HTML]{F2F2F2}0.51 &\cellcolor[HTML]{F2F2F2}100.79 \\
& es & 0.53 & 0.12 & 49.47 & 0.52 & 0.29 & 0.44 & 325.03 \\
& it & \cellcolor[HTML]{F2F2F2}0.48 &\cellcolor[HTML]{F2F2F2}0.10 &\cellcolor[HTML]{F2F2F2}48.08 &\cellcolor[HTML]{F2F2F2}0.48 &\cellcolor[HTML]{F2F2F2}0.25 &\cellcolor[HTML]{F2F2F2}0.41 &\cellcolor[HTML]{F2F2F2}124.49 \\
& total & 0.50 & 0.13 & 50.07 & 0.53 & 0.30 & 0.45 & 183.44 \\

\midrule

\multirow{4}{*}{WebNLG} & en & \,\,\cellcolor[HTML]{F2F2F2}0.55\textsuperscript{*} &\,\,\cellcolor[HTML]{F2F2F2}0.24\textsuperscript{*} &\cellcolor[HTML]{F2F2F2}57.05 &\,\,\cellcolor[HTML]{F2F2F2}0.72\textsuperscript{*} &\,\,\cellcolor[HTML]{F2F2F2}0.47\textsuperscript{*} &\,\,\cellcolor[HTML]{F2F2F2}0.57\textsuperscript{*} &\,\,\cellcolor[HTML]{F2F2F2}51.40\textsuperscript{*} \\
& es & \,\,0.56\textsuperscript{*} & \,\,0.17\textsuperscript{*} & 51.54 & \,\,0.63\textsuperscript{*} & \,\,0.38\textsuperscript{*} & \,\,0.49\textsuperscript{*} & \,\,150.84\textsuperscript{*} \\
& it & \,\,\cellcolor[HTML]{F2F2F2}0.55\textsuperscript{*} &\,\,\cellcolor[HTML]{F2F2F2}0.15\textsuperscript{*} &\cellcolor[HTML]{F2F2F2}51.75 &\,\,\cellcolor[HTML]{F2F2F2}0.59\textsuperscript{*} &\,\,\cellcolor[HTML]{F2F2F2}0.34\textsuperscript{*} &\,\,\cellcolor[HTML]{F2F2F2}0.47\textsuperscript{*} &\,\,\cellcolor[HTML]{F2F2F2}55.81\textsuperscript{*} \\
& total & \,\,0.55\textsuperscript{*} & \,\,0.19\textsuperscript{*} & 53.44 & \,\,0.64\textsuperscript{*} & \,\,0.40\textsuperscript{*} & \,\,0.51\textsuperscript{*} & \,\,86.02\textsuperscript{*} \\

\midrule
\midrule

\multicolumn{9}{c}{\textbf{Qwen-7B}} \\
\toprule
\multirow{4}{*}{Long-tail} & en & \,\,\cellcolor[HTML]{F2F2F2}0.65\textsuperscript{*} &\cellcolor[HTML]{F2F2F2}0.29 &\cellcolor[HTML]{F2F2F2}62.47 &\cellcolor[HTML]{F2F2F2}0.69 &\cellcolor[HTML]{F2F2F2}0.46 &\cellcolor[HTML]{F2F2F2}0.58 &\cellcolor[HTML]{F2F2F2}134.13 \\
& es & 0.60 & 0.16 & 53.43 & 0.57 & 0.34 & 0.49 & 181.75 \\
& it & \cellcolor[HTML]{F2F2F2}0.44 &\cellcolor[HTML]{F2F2F2}0.11 &\cellcolor[HTML]{F2F2F2}46.82 &\cellcolor[HTML]{F2F2F2}0.47 &\cellcolor[HTML]{F2F2F2}0.26 &\cellcolor[HTML]{F2F2F2}0.41 &\cellcolor[HTML]{F2F2F2}100.26 \\
& total & 0.56 & 0.19 & 54.24 & 0.58 & 0.35 & 0.49 & 138.71 \\

\midrule
\multirow{4}{*}{WebNLG} & en & \cellcolor[HTML]{F2F2F2}0.68 &\,\,\cellcolor[HTML]{F2F2F2}0.34\textsuperscript{*} &\cellcolor[HTML]{F2F2F2}66.33 &\,\,\cellcolor[HTML]{F2F2F2}0.79\textsuperscript{*} &\,\,\cellcolor[HTML]{F2F2F2}0.55\textsuperscript{*} &\,\,\cellcolor[HTML]{F2F2F2}0.64\textsuperscript{*} &\,\,\cellcolor[HTML]{F2F2F2}46.52\textsuperscript{*} \\
& es & 0.60 & 0.24 & 56.19 & \,\,0.66\textsuperscript{*} & \,\,0.43\textsuperscript{*} & \,\,0.53\textsuperscript{*} & \,\,117.41\textsuperscript{*} \\
& it & \cellcolor[HTML]{F2F2F2}0.44 &\cellcolor[HTML]{F2F2F2}0.15 &\cellcolor[HTML]{F2F2F2}46.96 &\,\,\cellcolor[HTML]{F2F2F2}0.54\textsuperscript{*} &\,\,\cellcolor[HTML]{F2F2F2}0.33\textsuperscript{*} &\,\,\cellcolor[HTML]{F2F2F2}0.45\textsuperscript{*} &\,\,\cellcolor[HTML]{F2F2F2}122.70\textsuperscript{*} \\
& total & 0.57 & 0.24 & 56.49 & \,\,0.66\textsuperscript{*} & \,\,0.44\textsuperscript{*} & \,\,0.54\textsuperscript{*} & \,\,95.54\textsuperscript{*} \\

\bottomrule
\end{tabular}
\caption{Performance comparison of models on long-tail triples from TailNLG and WebNLG test set. Asterisks ($^{*}$) denote a statistically significant difference with $p < 0.05$.}
\label{tab:web_vs_tail_full}
\end{table*}

\end{document}